\documentclass[10pt,twocolumn,letterpaper]{article}

\usepackage[pagenumbers]{cvpr}      %

\usepackage{graphicx}
\usepackage{amsmath}
\usepackage{amssymb}
\usepackage{booktabs}
\usepackage{bm}
\usepackage{bbm}
\usepackage{lipsum}

\usepackage{pifont}
\usepackage[pagebackref,breaklinks,colorlinks]{hyperref}

\usepackage[capitalize]{cleveref}
\crefname{section}{Sec.}{Secs.}
\Crefname{section}{Section}{Sections}
\Crefname{table}{Table}{Tables}
\crefname{table}{Tab.}{Tabs.}

\def\cvprPaperID{} %
\def\confName{}
\def\confYear{}

\DeclareMathOperator*{\concat}{||}

\begin{document}

\title{Road Extraction from Overhead Images with Graph Neural Networks }

\author{Gaetan Bahl\textsuperscript{1,3} \quad \quad Mehdi Bahri\textsuperscript{2} \quad \quad Florent Lafarge\textsuperscript{3}\\
\textsuperscript{1}IRT Saint Exupéry \quad \textsuperscript{2}Imperial College London \quad \textsuperscript{3}Université Côte d'Azur - Inria \\
{\tt\small \{gaetan.bahl,florent.lafarge\}@inria.fr,  m.bahri@imperial.ac.uk}
}

\maketitle

\begin{abstract}

Automatic road graph extraction from aerial and satellite images is a long-standing challenge. Existing algorithms are either based on pixel-level
segmentation followed by vectorization, or on iterative graph construction using next move prediction. Both of these strategies suffer from severe drawbacks, in particular high computing resources and incomplete outputs. By contrast, we propose a method that directly infers the final road graph in a single pass. The key idea consists in combining a Fully Convolutional Network in charge of locating points of interest such as intersections, dead ends and turns, and a Graph Neural Network which predicts links between these points. Such a strategy is more efficient than iterative methods and allows us to streamline the training process by removing the need for generation
of starting locations while keeping the training end-to-end. We evaluate our method against existing works on the popular RoadTracer dataset
and achieve competitive results. We also benchmark the speed of our method and show that it outperforms existing approaches. This opens the possibility of in-flight processing on embedded devices.

\end{abstract}

\section{Introduction}
\label{sec:intro}

\begin{figure}[t!]
    \centering
    \includegraphics[width=\columnwidth]{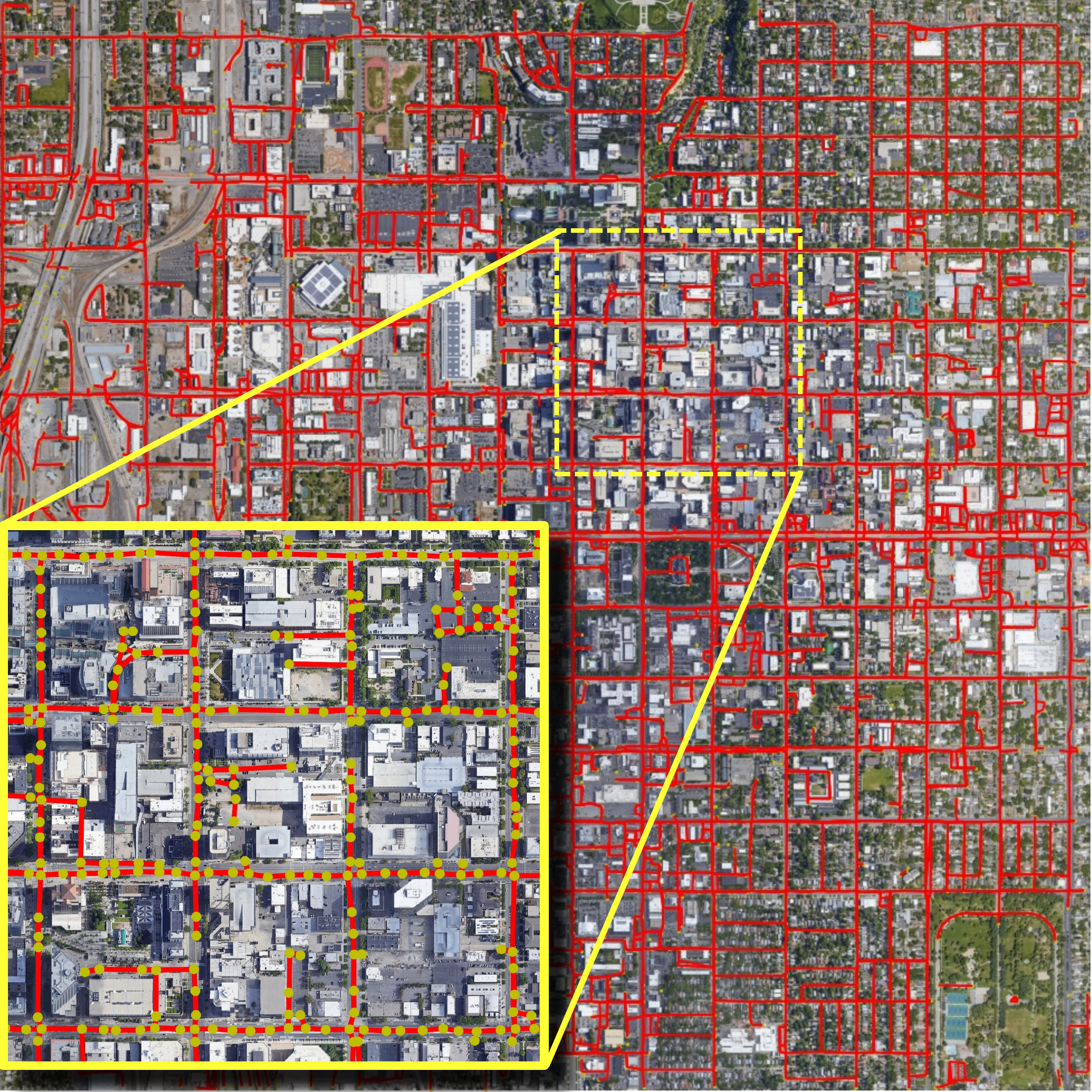}
    \caption{Result of our method on an image from the RoadTracer~\cite{Bastani2018} test set.
    Our method regresses an entire graph without the need of pre or post-processing.
    This 8192x8192 pixel image was processed in 22 seconds on a single GPU, which is up to \textbf{91 times faster}
    than previous approaches. The extracted graphs have a low complexity and retain good accuracy
    compared to existing works.}
    \vspace{-2em}
    \label{fig:teaser}
\end{figure}

Vector-based maps are heavily used in
Geographic Information Systems (GIS) such as online maps and
navigation systems.
The extraction of roads from overhead images has been, and 
mainly still is, performed by expert annotators. For a long time,
handcrafted methods have been developed
with the aim of reducing the burden of the annotators. However, they were not 
precise enough to replace them. Indeed, their outputs were imperfect
and had a high dependence on user-adjustable parameters \cite{Lian2020a}.
In recent years, the availability of large amounts of data \cite{russakovsky2015imagenet}, 
combined with the increase in compute capabilities, has led to a
rise in the popularity of deep learning methods based on 
Convolutional Neural Networks \cite{krizhevsky2012imagenet}.
Modern road extraction methods mostly fall into two kinds of 
approaches, each coming with its own drawbacks. 
The first kind \cite{Mattyus2017, zhang2018road, Zhou2018, Batra2019} relies on a pixel-based segmentation 
which is then converted to a graph using compute-intensive handcrafted algorithms that
require a lot of manual tuning and can often yield partial graphs.
The production of raster road masks using neural networks is a 
computationally expensive task in itself because of the large
number of layers required to retrieve a segmentation that has
the same resolution as the input image.
Iterative graph construction methods are
the second kind of deep learning based methods for road extraction \cite{Bastani2018, Li2019, Tan2020, Lian2020}.
These approaches explore the map from an initial starting location
using successive applications of a neural network on patches centered on the current point to predict the next-move.
While having the advantage of directly inferring a graph,
this is a very slow process. In addition, several starting locations
may need be picked in order to cover the possible multiple connected components 
of the final graph.

Our approach aims to solve the aforementioned issues of 
current deep learning based methods by offering a fast and convenient
way to extract roads from satellite or aerial images.
Inspired by recent advances in neural networks for object detection \cite{Tian2019a}, 
we design a fully convolutional detection head that learns to
precisely locate multiple points of interest
such as intersections, turns, and dead ends
across the whole image in a single pass. Node features attached to these
points are simultaneously regressed by this head, and fed to
a Graph Neural Network (GNN) which predicts links between these points 
to form the final graph without requiring any post-processing. 
These networks are jointly trained end-to-end from ground truth
road graphs, which are widely available from sources such as OpenStreetMap.
We benchmark our method against state-of-the-art approaches 
on the RoadTracer \cite{Bastani2018} dataset and find that it is 
orders of magnitude faster than recent competing methods while retaining
competitive accuracy. 
With the increasing interest in deep learning inference on embedded hardware,
such as satellites or drones \cite{bahl2019low},
our method, combined with recent advances in CNN and GNN quantization \cite{bulat2019xnor, bahri2021binary},
opens the possibility of on board in-flight road extraction, which could reduce
ground computation and bandwidth needs by transmitting graphs instead of images.

The main contribution of our work is threefold. First, we propose a fast fully convolutional method for precise localisation of 
    points of interest. Second, we design a Graph Neural Network for road link prediction. Third, we propose a performance-oriented architecture  highly competitive against state-of-the art
    approaches.

\section{Related Works}

\subsection{Convolutional Neural Networks}

In recent works, Convolutional Neural Networks (CNN)
such as VGG \cite{Simonyan2014} or ResNet \cite{He2015a} which were originally
used for image classification, are stripped of their final classifier
and repurposed as fully convolutional feature extractors (often called backbones) for other
computer vision tasks, such as semantic segmentation \cite{Long2014}.
Object detection methods such as YOLO \cite{Redmon2015} or FCOS \cite{Tian2019a}
have taken advantage of the fully convolutional nature of 
these backbones to design very fast
single stage object detection neural networks, which infer bounding boxes
and classes in a single forward pass and can be trained end-to-end, 
as opposed to two-stage approaches that perform these tasks separately.
Single-stage architectures often follow a similar structure 
composed of a backbone which extracts features, a neck outputting
feature representations at different scales \cite{Lin2016}, 
and several detection heads performing the final task.

\subsection{Road Extraction} 
A large number of unsupervised methods have been developed for road graphs extraction. 
We refer the reader to recent extensive surveys on the subject for more information \cite{Lian2020}. 
In this section, we focus on recent road extraction methods based on deep learning. 
Early deep learning based methods for road extraction
were patch-based applications of fully connected 
neural networks \cite{Mnih2010}.
However, most recent neural network-based methods  
proceed differently and
can be sorted into two groups. 

The first group of methods is based on pixel-level segmentation.
Inspired by the success of FCN \cite{Long2014} and U-Net \cite{Ronneberger2015},
these methods use a backbone as an encoder and 
learned upscaling as a decoder to retrieve
a classification of individual pixels of an input image.
In \cite{zhang2018road}, residual connections are added to a U-Net
to improve road predictions.
D-LinkNet \cite{Zhou2018} adapts the LinkNet architecture to form
a U-Net-like network and achieve better connectivity.
DeepRoadMapper \cite{Mattyus2017} use a ResNet-based\cite{He2015a} FCN to produce a segmentation
which is then converted to a graph using thinning and an additional 
connection classifier.
In \cite{Batra2019}, the authors improve the connectivity of extracted networks
by jointly learning the segmentation and orientation of roads.
All segmentation-based approaches have the drawback of relying on post-processing
techniques to convert the predicted road pixels to a road graph. Thus, these methods
are slow and the final graphs can lack connectivity. 

The second kind of approaches iteratively construct a graph
from an initial starting point
by successive applications of a neural network for next move 
prediction around the center of an input patch. These methods
are inspired by the way human annotators iteratively create road graphs.
In \cite{Ventura2018}, a network predicts which points from
the border of the output patch are linked to the center point.
RoadTracer \cite{Bastani2018} uses a CNN to predict the direction
of the next move with a fixed step size.
PolyMapper \cite{Li2019} uses a Recurrent Neural Network (RNN)
guided by segmentation cues to predict the road topology and building
polygons.
VecRoad \cite{Tan2020} implements a variable step size for next move 
prediction in order to achieve better alignment of intersections.
DeepWindow \cite{Lian2020} estimates the road direction along
a center line regressed by a CNN
using a Fourier spectrum analysis algorithm.
While these methods have the advantage of outputting a road graph
directly, they cannot guarantee the exploration of the whole map
from a single starting location. The choice of initial starting locations
is a hard problem in itself.
Moreover, the repeated application of CNNs makes these approaches very slow.

\begin{figure*}[t!]
    \centering
    \includegraphics[trim={2mm 1mm 1mm 5mm}, width=\textwidth]{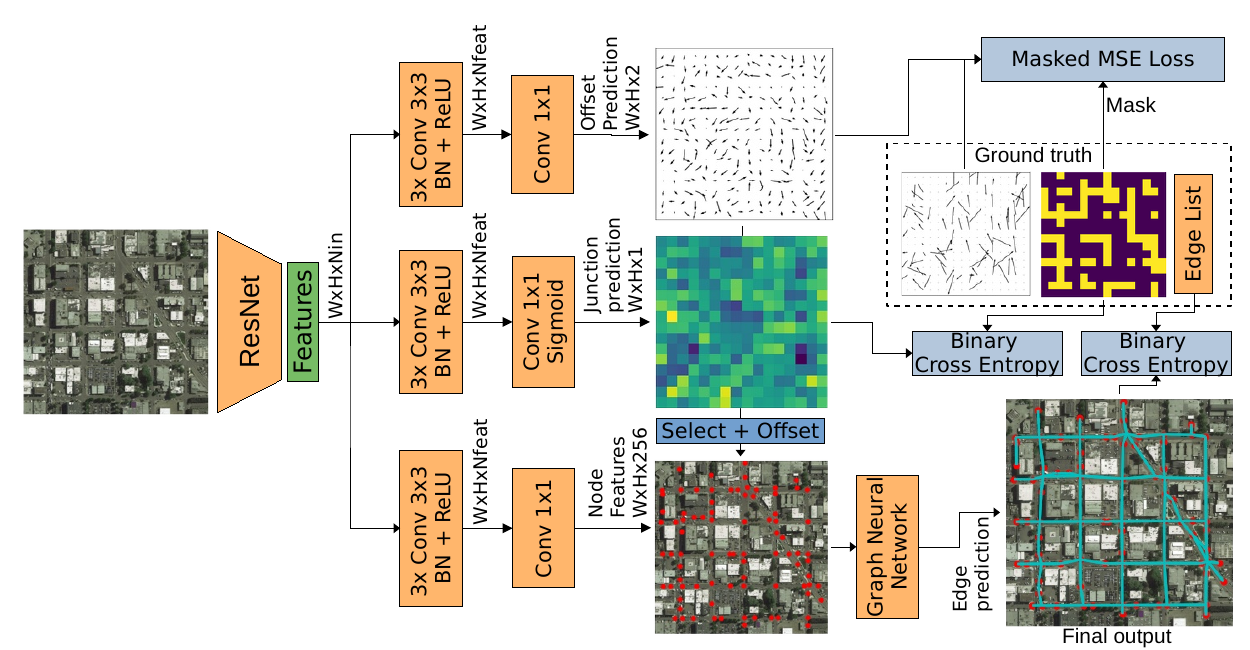}
    \caption{Our single-shot road graph extraction neural network. Features computed using a ResNet \cite{He2015a} backbone
    are fed to three branches. The junction prediction branch identifies cells which contain points of interest and is trained
    with binary cross-entropy. 
    The offset prediction branch regresses the precise location of these points inside of the cells by outputting a two-dimensional
    vector field and is trained with a Mean Squared Error loss.
    Points selected by the junction and offset branch are paired with node features from a third branch and fed
    to a Graph Neural Network which predicts the presence of edges between pairs of points to create the final output.
    }
    \label{fig:method}
\end{figure*}

\subsection{Relational Inference and GNNs} Graph Neural Networks (GNNs) \cite{Scarselli2009} are effective models for encoding interactions, and have recently been applied to link prediction \cite{NEURIPS2018_53f0d7c5}, relational structure discovery, such as to model the dynamics of complex systems \cite{kipf2018neural}, and to the modeling of relational knowledge \cite{Schlichtkrull}. Kipf and Welling \cite{kipf_variational_2016} proposed a graph autoencoder using a Graph Convolutional Network (GCN) encoder \cite{Kipf2017} and a simple dot-product decoder to model undirected networks and applied it to the link prediction task. Follow-up work \cite{Schlichtkrull} models different edge types in relational data using a graph neural network encoder and a decoder based on the DistMul \cite{Yang2014DistMul} factorization. In \cite{kipf2018neural} the authors propose a message-passing \cite{Gilmer2017} encoder that alternates between the computation of node and edge features that can later be leveraged to predict the evolution of the system of interest several time steps in advance. The aforementioned models assume the ground-truth graph is partially known, or operate on the complete graph as an uninformative prior \cite{kipf2018neural}. A parallel line of work is that of learning the unknown graph explicitly. The Dynamic Graph CNN model \cite{Wang2018_dgcnn} proposes to dynamically build a graph by \textit{k}-NN search in the feature space after each application of the EdgeConv operator also introduced in \cite{Wang2018_dgcnn}. The approach has been extended in \cite{Kazi2020}, where the authors address the question of the differentiable construction of a discrete graph using the recently proposed Gumbel Top-\textit{k} trick \cite{Kool2019StochasticBA} - a stochastic and differentiable relaxation of \textit{k}-NN search - and decouple the construction of the graph and the learning of graph features amenable for downstream tasks. With the introduction of deeper architectures \cite{Li2019DeepGCN, Gong2020} came increased interest for efficient implementations of GNNs, an issue relevant to our work as on-board processing of satellite images, such as for road graph extraction, is a pressing issue. \cite{Wang2020} applied bucket-based quantization of matrix-matrix products to accelerate the GCN \cite{Kipf2017} operator. \cite{bahri2021binary} proposed a general framework for binarizing graph neural networks and specifically introduced efficient binarized versions of the Dynamic EdgeConv operator \cite{Wang2018_dgcnn} with real-world speed-ups on a low-power device.

\section{Method}

In this section, we present our method based on a fully convolutional
head for regression of points of interest and a Graph Neural Network (GNN) for link prediction.
An overview of our method is available in Figure \ref{fig:method}.

\subsection{General architecture}

Our neural network architecture follows the trend of using a pre-trained
CNN as a feature extractor (backbone). Most detection neural networks also
use a Feature Pyramid Network (FPN) as a "neck" that aggregates features from 
multiple scales. However, we chose not to use an FPN since satellite images
have a fixed ground sampling distance and consequently, models that
work on them are less vulnerable to changes in scale. Since we work at a
single level of detail, we feed the output feature maps of the feature 
extractor directly to a single head. These changes allow us to save 
computation time and make the training process more streamlined. Indeed, 
we remove all ground truth bounding box assignment complications that arise
when using multiple heads.

Each branch of our head is composed of three convolutional layers
working on $N_{feat}$ feature maps. The first two branches are the junction-ness
and offset branch, which form the point-of-interest detection model. 
The third branch is a node feature branch, which computes features that will be used
to predict links between points detected by the "point-of-interest" branch.

Let $H_I,W_I$ be the dimensions of the input RGB image $I \in \mathbb{R}^{H_I\times W_I \times 3}$.
With the ResNet-50 \cite{He2015a} backbone used in our experiments, 
the height and width $H,W$ of the input feature 
maps $F^{in} \in \mathbb{R}^{H\times W \times N_{in}}$ 
are 32 times smaller than the ones of $I$. 
This means that the final layers of our head have $H \times W$ output cells.
In the case of an input image size of $512\times 512$ pixels, for example,
we thus obtain 256 output cells, which can each regress the position of one point.

\subsection{Point detection branch}

Our point of interest detection branch, shown in Figure \ref{fig:method} is inspired
by recent developments in single shot detection 
neural networks \cite{Tian2019a, Redmon2015}, which regress a bounding box for each output
"cell" of a detection head, even if they do not necessarily contain 
an object. 
This allows object detection to be performed using very fast Fully Convolutional 
Networks. 
In \cite{Tian2019a}, the notion of "center-ness" is then introduced
to filter out cells that do not belong to any object.
These methods represent bounding boxes as four offsets relative to the center of the cell. 
We adapt this design to the regression of points of interest
by outputting a "junction-ness" score $J_{j}$ for each output cell $j$, as well as 
a two-dimensional vector $\mathbf{o}_{j} = [u_{j},v_{j}] \in [-0.5;0.5]^2$ representing
the offset of the regressed point with respect to the center of its cell.
A point $\mathbf{p}_j$ is detected in an output cell $j$ if $J_{j} > J_{thr}$, where $J_{thr}$ is
the junction-ness detection threshold. We can retrieve the 
coordinates $(x_j, y_j)$ of $\mathbf{p}_j$ in the original input image:

\begin{equation}
    x_j = \frac{u_{j}  + X_j + 0.5}{W} \cdot W_I , 
    y_j = \frac{v_{j}  + Y_j + 0.5}{H} \cdot H_I
\end{equation}

where $(X_j, Y_j)$ are the coordinates of the cell $j$ in the output feature map $F^{out}$.

\begin{figure}[h!]
    \centering
    \includegraphics[width=0.9\columnwidth]{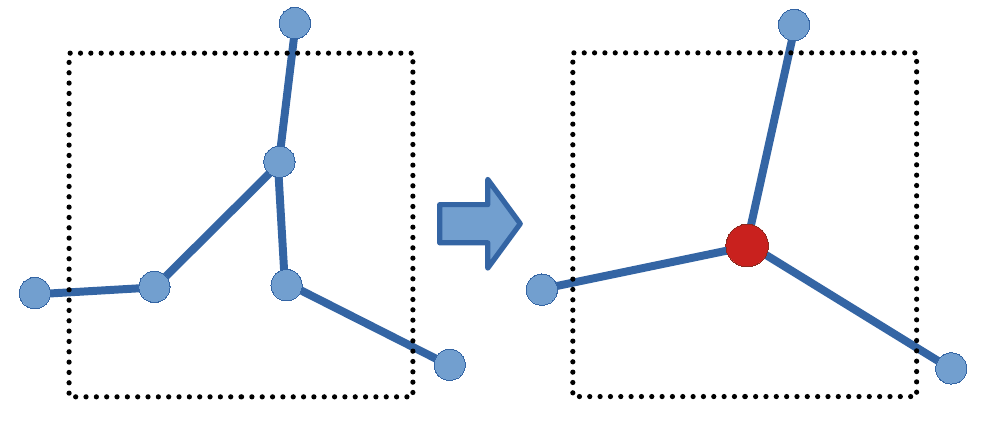}
    \vspace{-1em}
    \caption{Node averaging which occurs when several ground truth points
    belong to the same output cell. The offset branch is thus trained to regress
    the centroid of these ground truth points.}
    \label{fig:collapse}
\end{figure}

\subsection{Choice of input resolution}

\begin{figure}[h!]
    \centering
    \includegraphics[width=0.9\columnwidth]{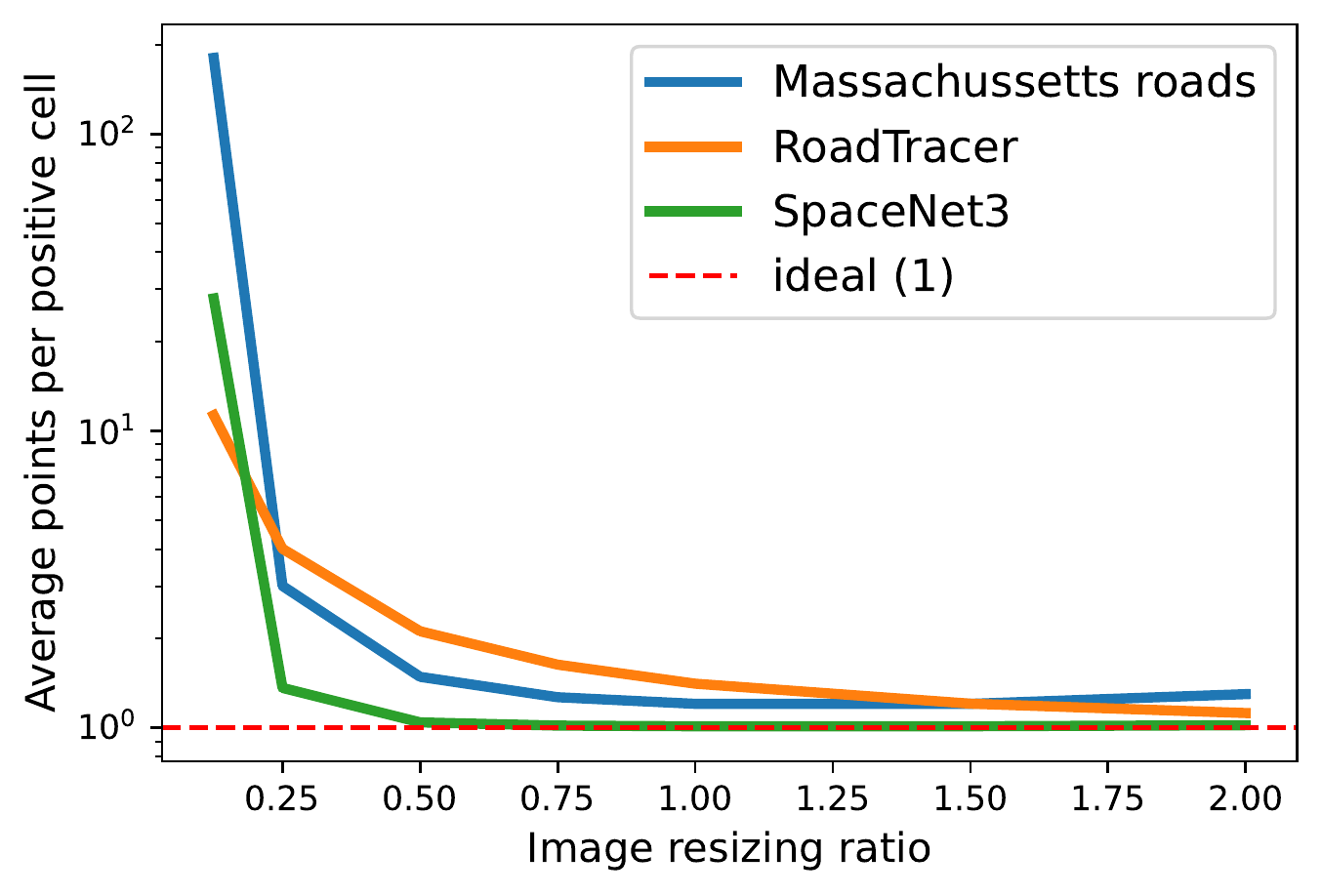}
    \caption{Average ground truth points per positive output cell at a range of image resizing ratios
    for 3 popular road extraction datasets. The ideal average is one, to prevent the collapse of 
    neighboring nodes as much as possible.}
    \label{fig:ratio}
\end{figure}

By design, our method is only able to find a single point
of interest by output cell and is directly trained on ground
truth annotations. 
However, depending on the dataset,
and especially in very dense neighborhoods,
 several ground truth points may fall
into the same cell.  We chose that 
these points would effectively be merged into their centroid, while
keeping incoming edges from outside of that cell linked to that new point,
as shown on Figure \ref{fig:collapse}.
However, this situation should be avoided as much as possible, 
since it can shift intersections, and link close points that should not be linked (e.g. parallel roads).
Images are often resized when being forwarded through a neural network, as
a compromise between speed, memory and accuracy.
In our case, the resizing ratio has to be chosen carefully according
to the ground sampling distance (GSD) of the dataset and the density
of ground truth annotations.
To estimate the correct ratio for each dataset, we evaluate 
the average number of points in each positive cell over the whole
dataset, at a wide range of ratios. This average should be as close to one 
as possible, in order to limit the effects of point merging.
Figure \ref{fig:ratio} shows this average for the popular
RoadTracer \cite{Bastani2018}, SpaceNet3 \cite{van2018spacenet}
and Massachussetts roads \cite{mnih2013machine} datasets.

\subsection{Edge prediction}

Once junctions have been extracted from the image, our task is to predict which ones are connected. We cast the problem as that of learning a latent graph. Starting with an initial prior connectivity estimation - \ie a set of edges - $\mathcal{E}^0$, we aim at both inferring missing edges and discarding irrelevant ones.

Formally, let $j$ be one of the detected junctions. We denote by $X_j = F^{out}_j$ the corresponding features in the output feature map of the backbone, and $\mathbf{p}_j = (x_j, y_j)$ the 2D Cartesian coordinates of the junction in the input image, computed from the offset vectors.

Our method computes initial node embeddings 
\begin{equation}
\mathbf{x}_j = f(X_j, \mathbf{p}_j)    .
\end{equation}
We choose the function $f$ to be either a 2D Convolutional Neural Network defining an additional node features branch of the network - responsible for dimensionality reduction and transformation of the raw image features - or the identity, the latter case leaving the task of transforming the raw image features to the GNN. We then perform several (\eg three) message-passing iterations using the EdgeConv operator \cite{Wang2018_dgcnn}:
\begin{align}
    \mathbf{e}^{(l)}_{ij} & = \text{ReLU} \left( \bm{\theta}^{(l)} (\mathbf{x}^{(l-1)}_j - \mathbf{x}^{(l-1)}_i) + \bm{\phi}^{(l)} \mathbf{x}^{(l-1)}_i \right)\\
              & = \text{ReLU} \left( \bm{\Theta}^{(l)} \tilde{\mathbf{X}}^{(l-1)} \right)\\ \label{eq:edgeconv}
    \mathbf{x}^{(l)}_{i} &= \max_{j \in \mathcal{N}(i)} \mathbf{e}^{(l)}_{ij}
\end{align} followed by Batch Normalization, where $\tilde{\mathbf{X}}^{(l-1)} = \left[ \mathbf{x}^{(l-1)}_i \concat \mathbf{x}^{(l-1)}_j - \mathbf{x}^{(l-1)}_i \right]$, $\bm{\theta}$ and $\bm{\phi}$ are trainable weights, and $\bm{\Theta}$ their concatenation. The resulting increase in receptive field allows for connectivity patterns to be learned based on both local image and graph properties, and shared context across junctions.

\begin{figure}
    \centering
    \includegraphics[width=.9\linewidth]{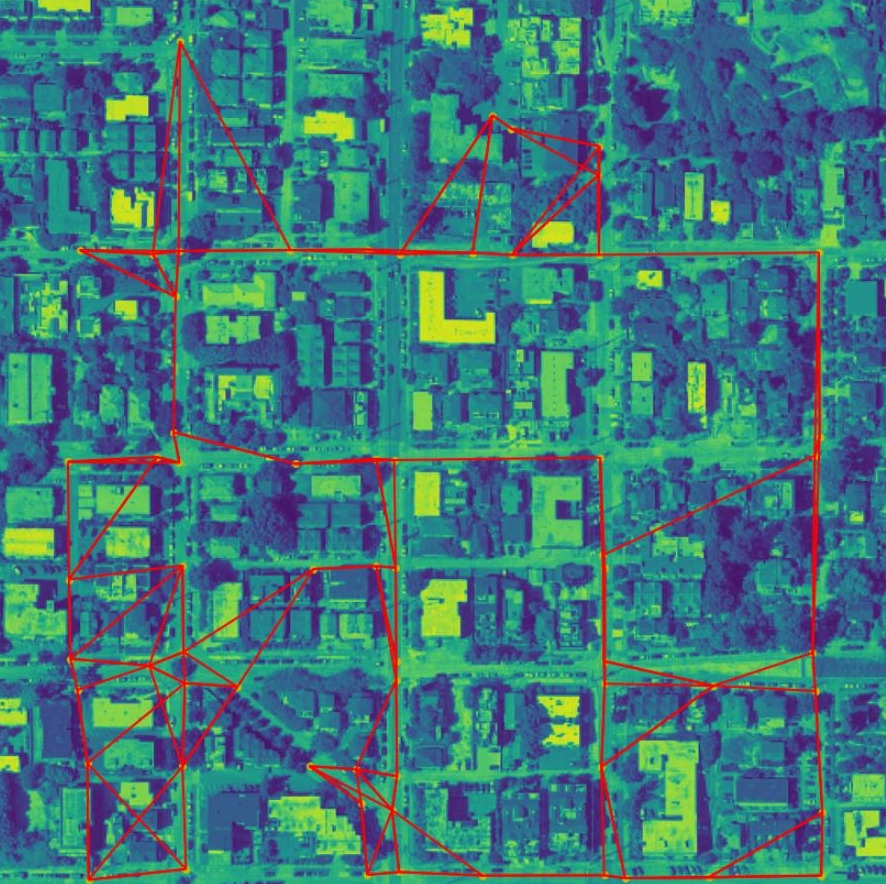}
    \caption{$k = 4$-NN graph constructed in feature space on features learned by an additional \textit{node features head} combined with an MLP edge classifier.}
    \label{fig:knn_init}
\end{figure}

Finally, each possible edge is scored using a simple scoring function $g$ applied on the final node features of the graph. While we could opt to classify the edges based on the computed edge features $\mathbf{e}_{ij}$, performing message passing on a sparse graph (potentially built dynamically) is desirable for efficiency at both training and inference. We note
\begin{equation}
    p_{ij} = \sigma(g(\mathbf{x_i, x_j}))
\end{equation} the inferred probability that the edge $\mathbf{e}_{ij}$ exists, where $\sigma$ is the sigmoid function. In practice, we choose $g$ to be either a single bilinear layer, \ie
\begin{equation}
    g(\mathbf{x}_i, \mathbf{x}_j) = \mathbf{x}_i^T W_g \mathbf{x}_j
\end{equation} or a multi-layer MLP classifier (please refer to the appendix for more details and comparison with other decoders such as the dot product).

\paragraph{Connectivity estimation} Our choice of operator is motivated by the fact the connectivity is unknown, and therefore we may choose $\mathcal{E}^0$ to the set of edges of the complete graph (\ie considering all possible connections between junctions), or introduce a sparse prior by building the graph dynamically by $k$-NN search in feature space. We also considered the case where the initial graph is complete, while subsequent iterations of message passing are done on dynamically inferred connectivity. Choosing $k = 4$ is motivated by the observations that junctions with more than four incident roads are rare. To verify this intuition, we trained a model using a three-layer (2D convolutions followed by Batch Normalization and ReLU activations) node features head and an MLP classifier for each possible edge. The model was trained to convergence, and the four nearest neighbours graph was built on the output of the node features branch. The resulting graph is shown in Figure \ref{fig:knn_init}, and demonstrates that a high quality sparse initialization of the connectivity can be obtained in this manner.

\newcommand{\loss}[1]{\mathcal{L}_\text{#1}}
\newcommand{\lambdaloss}[1]{\lambda_\text{#1}}
\newcommand{\Ntxt}[1]{N_\text{#1}}

\begin{figure*}[t!]
    \centering
     \setlength\tabcolsep{2pt}
    \begin{tabular}{ccc}
        \includegraphics[width=0.33\textwidth]{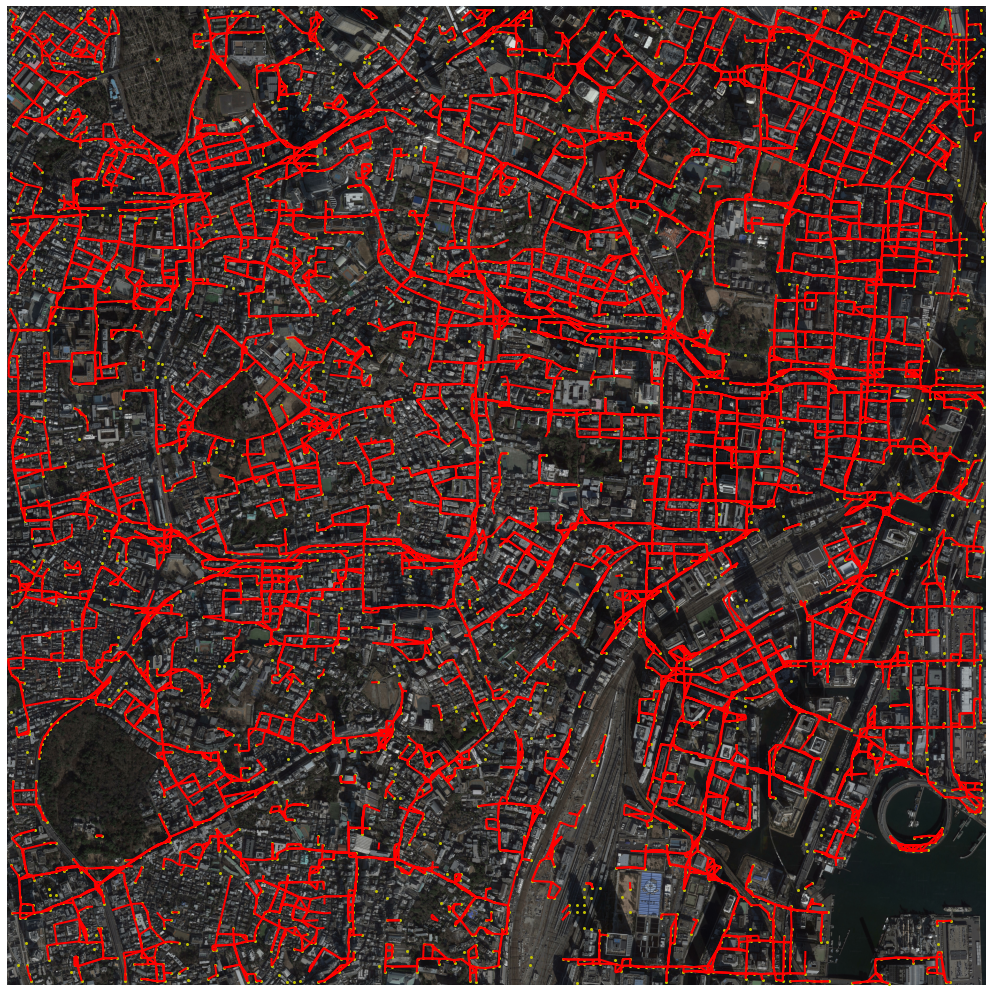}&
        \includegraphics[width=0.33\textwidth]{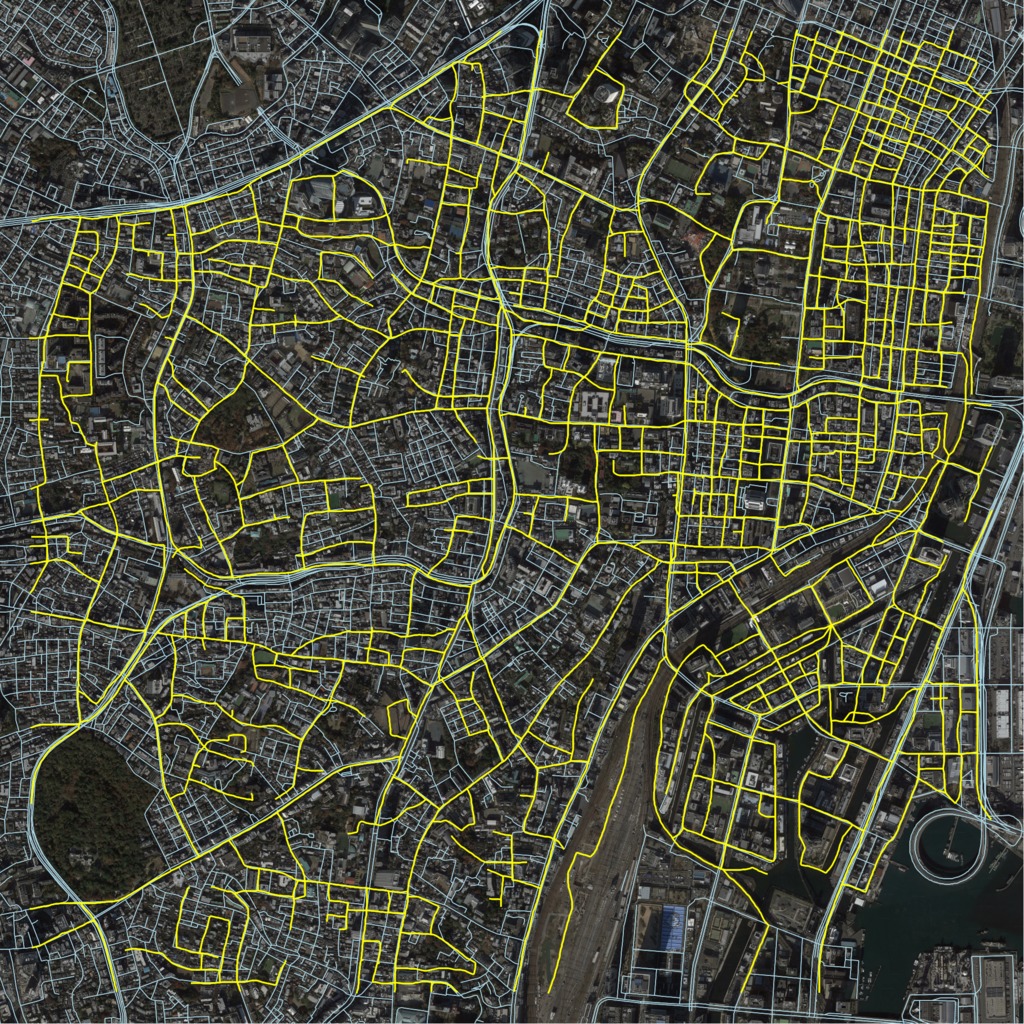}&
        \includegraphics[width=0.33\textwidth]{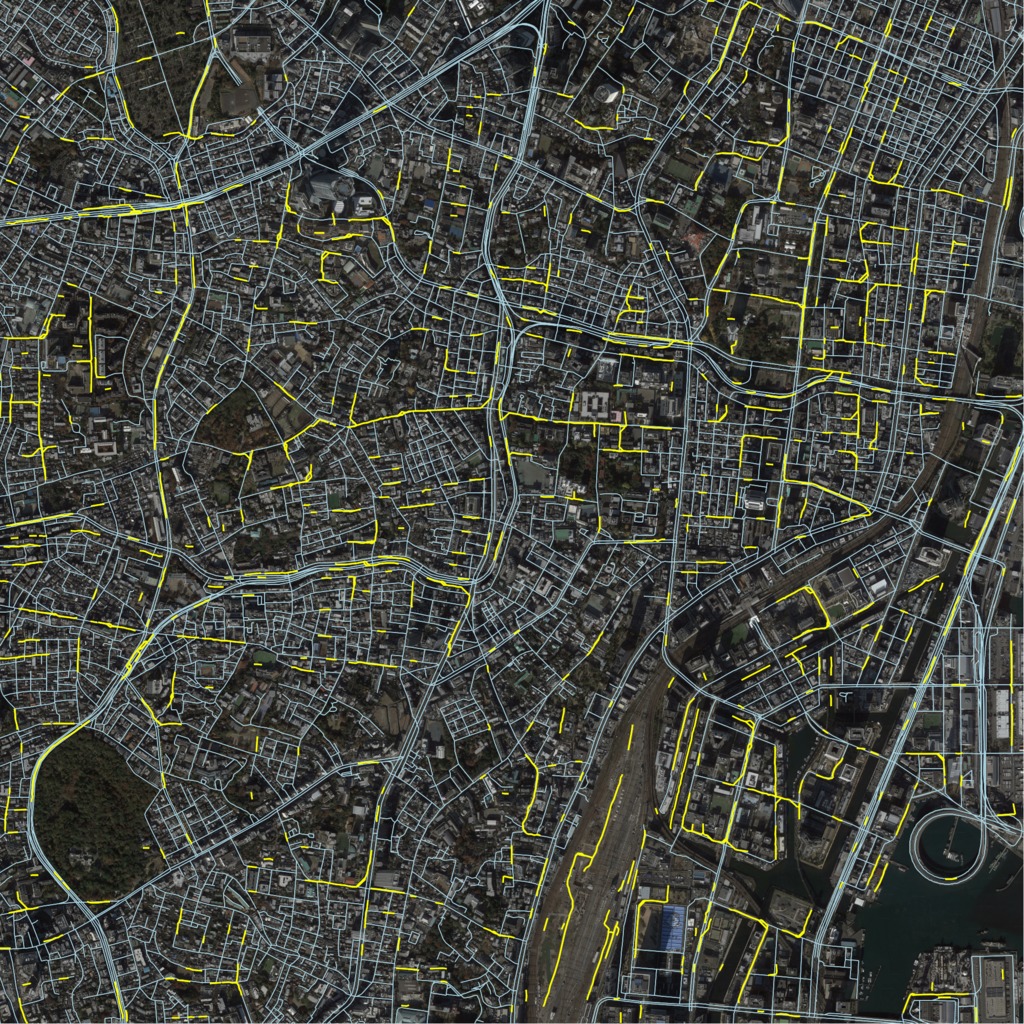}
    \end{tabular}
    \caption{Comparison of our method (left) against an iterative method \cite{Bastani2018} (middle) and a segmentation based method \cite{Mattyus2017} (right). Our method is able to find more roads in this challenging area of the RoadTracer test set, even 
    in sections that are completely missed by other methods.}
    \label{fig:qualitative}
\end{figure*}

\subsection{Loss function}

To train our neural network our loss function is defined as a sum of three terms:
\begin{equation}
    \mathcal{L} = \loss{jun} + \loss{off} + \loss{edge}
\end{equation}
where $\loss{jun}$ and $\loss{edge}$ are the 
binary cross-entropy loss
which we use to train our junction-ness branch and our edge prediction Graph Neural Network. 

$\loss{off}$ is a Mean Squared Error offset loss masked with the positive junction predictions, which 
we use to train our offset regression branch. We define it as:

\begin{equation}
    \displaystyle \loss{off} = \frac{1}{\sum \mathbbm{1}_{J_{i,j} > J_{thr}} } \sum_{i,j} \mathbbm{1}_{J_{i,j} > J_{thr}} (v_{i,j} - V_{i,j})^2
\end{equation}

Where $V$ is the ground truth offset vector field. The mask
is used so that the predictions of the offset branch
are not conditionned to the presence of nodes.

\subsection{Inference on large images}

In our testing, we found that the output of the link prediction
branch is tied to the original training resolution of the network.
Thus, we did not obtain good results by directly inferring on a 
bigger image. Instead, we apply our network on large images
using a sliding window of the same size as
the training resolution, and accumulate the detected edges.

\section{Experiments}

In this section, we 
 compare our method against other road extraction approaches on
 the widely-used RoadTracer dataset \cite{Bastani2018},
which is composed of 35 training cities (180 images of 4096x4096 pixels)
and 15 test cities of 8192x8192 pixels. The low number of training images,
low ground sampling distance, high resolution and high annotation density
of this dataset make it quite challenging.

\subsection{Implementation details}

According to the results shown
in Figure \ref{fig:ratio}, we choose a rescaling ratio of 0.5 as a good
compromise between accuracy and speed.
We implement our method in the Pytorch 1.9.1 \cite{NEURIPS2019_9015}
framework based on CUDA 11. We choose a ResNet-50 backbone (thus
$N_{in} = 2048$) and $N_{feat} = 256$. We perform edge classification on the complete graph (other cases included in the appendix) using a three-layer GNN and a 2-layer MLP scoring function. We use basic data augmentation techniques
such as random flips, and train our network on 512x512 pixel random crops
for 2350 epochs, which took 24 hours on our system with a single Nvidia RTX 3090
GPU with 24GB of VRAM, an AMD Ryzen 9 3900X CPU and 64GB of RAM. We use the
Adam \cite{kingma2014adam} optimizer with a learning rate of $1e^{-3}$.

We also run performance benchmarks.
All numbers were obtained on a computer with two Intel Cascade Lake 6248 20-core CPUs, 192GB of RAM,
and four Nvidia Tesla V100 GPU with 16GB of VRAM. 10 CPU cores, a single GPU, 
and a quarter of the available memory were assigned 
to each run.

\subsection{Qualitative results}

Figure \ref{fig:qualitative} shows qualitative results against an
iterative method and a segmentation-based method. Our method is able to 
find a noticeably larger number of roads than RoadTracer \cite{Bastani2018} 
and DeepRoadMapper \cite{Mattyus2017} in this challenging image.
We can also notice that there are a few areas with low connectivity compared
to RoadTracer. Indeed, since their network "walks" through the entire regressed graph, 
it is inherently made of a single connected component. 
However, the result of RoadTracer is missing entire portions of the city (e.g. top left)
because they are separated from the starting position by a highway that the network was
not able to cross during its exploration. Our method does not suffer from such limitations.
More qualitative results are available in Figure \ref{fig:teaser} and in appendix.

\subsection{Quantitative evaluation}

We follow the literature and use the three
metrics defined in \cite{Tan2020} to evaluate the accuracy
of our network. P-F1 is a pixel-based F1 score obtained by
comparing the rasterized output graph and ground truth. J-F1 is
a junction-based F1 score based on local connectivity. APLS
is the Average Path Length Similarity defined in the SpaceNet
challenge \cite{van2018spacenet} and is based on the comparison
of shortest paths in the predicted graph and the ground truth.

The results of our testing are shown in Table \ref{tab:accuracy}.
Our method achieves results which are competitive with DeepRoadMapper \cite{Mattyus2017}
and RoadTracer \cite{Bastani2018}.
It is however outperformed by VecRoad \cite{Tan2020}. Our intuition is that VecRoad
sort of "brute-forces" these metrics by regressing a lot more points
and edges than our method, as shown in Section \ref{graph_complexity} and Table \ref{tab:complexity}.

\begin{table}[h!]
    \centering
    \begin{tabular}{c|c|c|c}
    Method & P-F1 & J-F1  & APLS \\ \hline
    DeepRoadMapper \cite{Mattyus2017} & 56.85  & 29.05 & 21.27 \\
    RoadTracer \cite{Bastani2018}&  55.81 & 49.57 & 45.09\\
    RoadCNN\cite{Bastani2018} & 71.76  & --  & --  \\
    VecRoad\cite{Tan2020} & \textbf{72.56} & \textbf{63.13} & \textbf{64.59} \\\hline
    Ours & 57.2 & 39.23  & 46.93 \\
    \end{tabular}
    \caption{Accuracy metrics on the RoadTracer Dataset, evaluated using open source code from \cite{Tan2020} and \cite{Bastani2018}.
    RoadCNN is a purely segmentation-based method implemented by \cite{Bastani2018}, thus it only has a P-F1 score.}
    \label{tab:accuracy}
\end{table}

\subsection{Performance benchmarks}

Run-time is seldom mentioned
in the road extraction literature, but it is a very important metric,
especially as we move towards inference on edges devices.
Thus, we benchmark the performance of our method against other recent approaches
with open source code.
 We remove data loading and output times from all methods
and thus only account for pre-processing, inference and post-processing steps
of each algorithm. We provide average performance numbers for a single 8192x8192 test
image from the RoadTracer \cite{Bastani2018} dataset, as well as numbers
for the whole dataset (15 images), since some 
methods such as VecRoad \cite{Tan2020} have optimizations for
multiple-image inference. Our method can run on the whole dataset at once
by performing the sliding window in batches, which is a bit faster.

\begin{table}[h!]
    \centering
    \small
        \setlength\tabcolsep{3pt}
    \begin{tabular}{c|c|c|c}
    Method & Type & Single image & Whole dataset \\ \hline
    VecRoad \cite{Tan2020} & Iterative & 1902.5  & 8873.4 \\
    RoadTracer \cite{Bastani2018} & Iterative & 579.4 & 8690.5* \\\hline
    DeepRoadMapper \cite{Mattyus2017} & Seg. & 1361.6  & 20423.4 \\
    RoadCNN \cite{Bastani2018}& Seg. & 58.6 & 878.5 \\ \hline
    Ours & Graph & \textbf{20.9} & \textbf{295.0}\\
    \end{tabular}
    \caption{Run-time in seconds on the RoadTracer Dataset.
    The single image score is averaged over the whole test set. 
    Our method is the fastest by a large margin. *RoadTracer failed
    on some images, thus this number is extrapolated from the average.}
    \label{tab:benchmark}
\end{table}

\begin{figure}[h!]
    \centering
    \includegraphics[width=0.9\columnwidth]{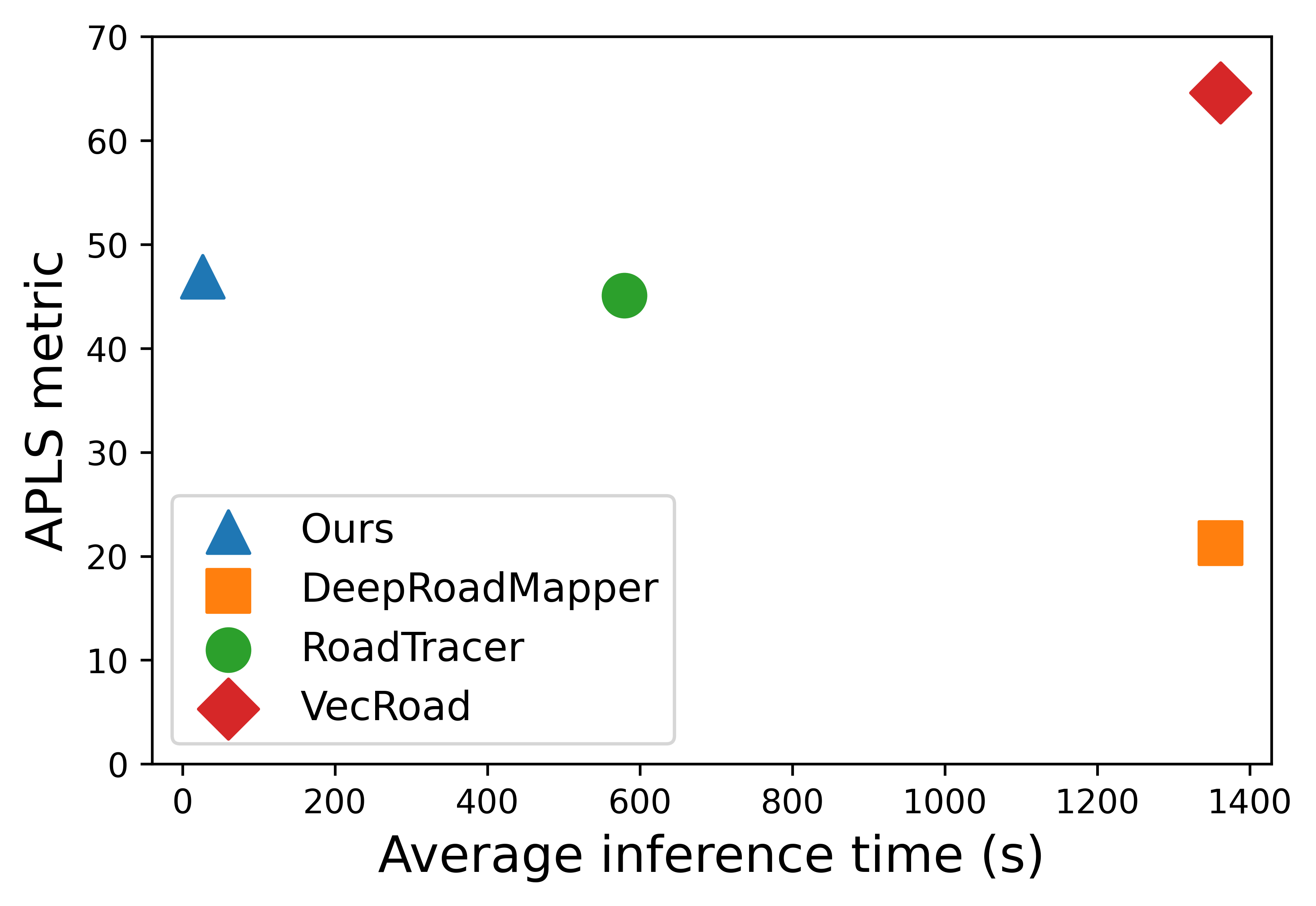}
    \caption{APLS metric with respect to inference time for our method, 
    compared to state-of-the art approaches. Our method is the fastest and offers a very good compromise
    between accuracy and speed.}
    \label{fig:speed}
\end{figure}

The results of this experiment are shown in Table \ref{tab:benchmark} and Figure \ref{fig:speed}.
We observe that iterative methods are much slower because of the 
repeated forward passes required for next move estimation. 
Segmentation methods are not particularly fast as well, as 
the upscaling computation performed by a learned decoder is very compute intensive,
and the extra post-processing steps required for graph conversion
are slow.
Our method is the fastest one by a large margin and offers a very good compromise
between accuracy and speed. Indeed, it achieves similar APLS scores as RoadTracer \cite{Bastani2018}
while running 27 times faster. Our approach also beats the APLS scores of DeepRoadMapper \cite{Mattyus2017}
by a large margin while running 65 times faster. 
It has lower accuracy scores than VecRoad \cite{Tan2020}, however
it is 91 times faster. 
Our method could be made even faster by using a smaller and more efficient backbone,
such as ResNet-18 \cite{He2015a} or DarkNet-53 \cite{Redmon2015}.

\subsection{Graph complexity}

\label{graph_complexity}

Another interesting metric which is seldom
mentioned in the road extraction literature is the
notion of graph complexity. Graphs are a sparse 
representation and are meant to use a lot less storage space than
masks. 
Most algorithms commonly used on graphs (e.g. shortest path algorithms)
have a complexity which depends on the number of nodes and edges of the graph.
Thus, offering accurate graphs with a low number of 
unnecessary nodes and edges is also beneficial in terms of run-time of 
subsequent applications.
To evaluate the complexity of the graphs regressed by our method and
compare it to other approaches, we propose a complexity score (lower is better)
which is simply the total number of graph elements (nodes and edges)
divided by the APLS score. We use the APLS because it seems to be the most popular metric. 
This score represents the compromise between the accuracy and compactness of graphs.

\begin{figure}[h!]
    \centering
    \includegraphics[width=0.49\columnwidth]{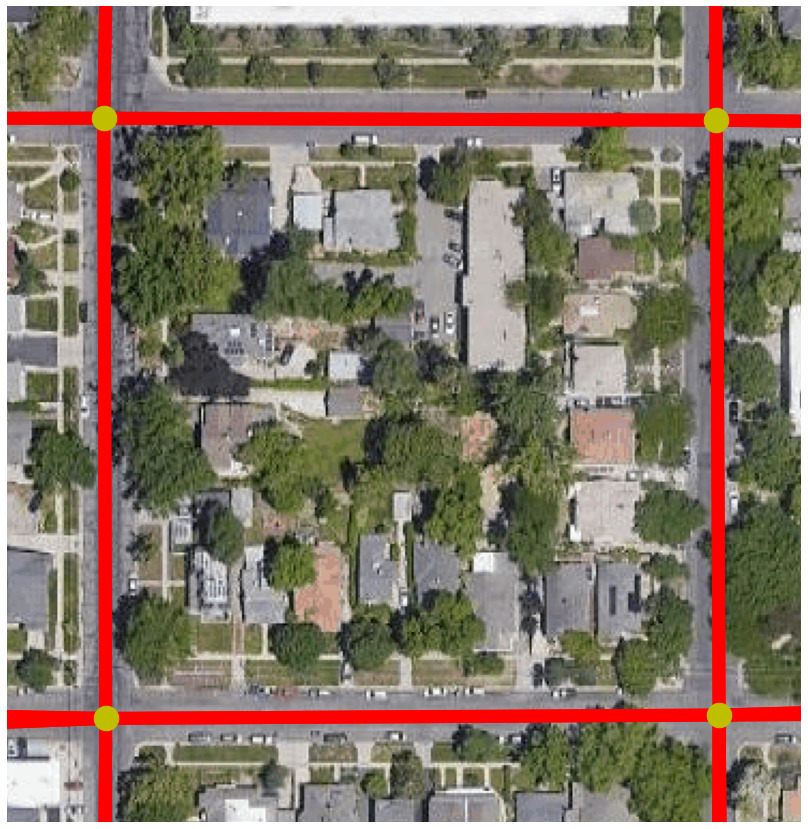}
    \includegraphics[width=0.49\columnwidth]{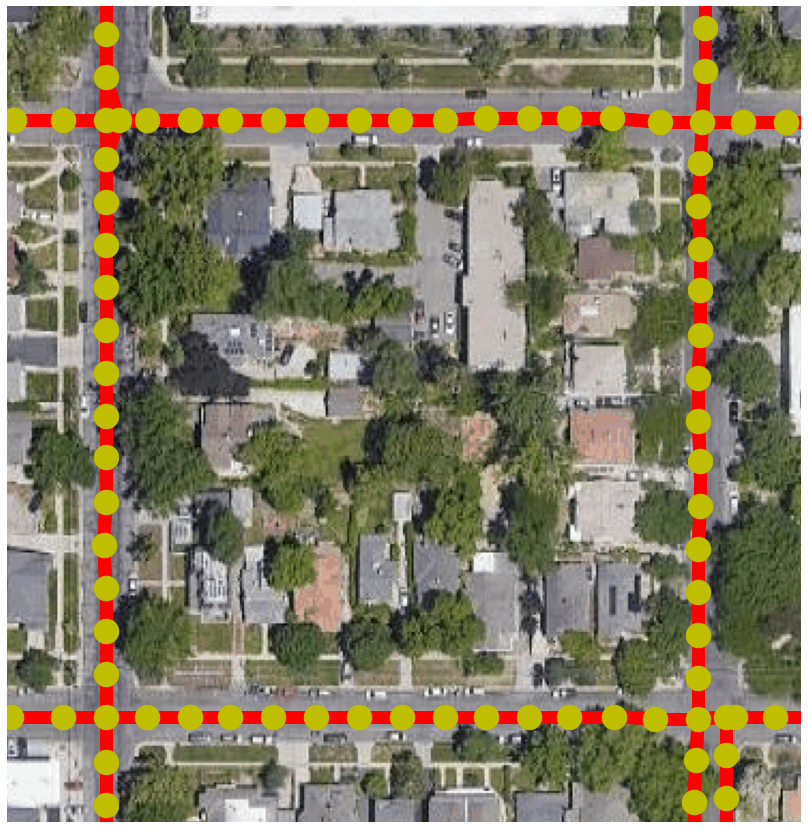}
    \caption{ Exemple of the low complexity of the graphs inferred by \textbf{our method (left)}
    compared to VecRoad \cite{Tan2020} (right). Our method finds an optimal 4-node
    representation of this simple neighborhood, whereas iterative methods tend to find
    overly complex representations. Having graphs with a low number of elements will 
    save storage and
    speed up the computation of algorithms performed on these graphs, such 
    as shortest path algorithms.}
    \vspace{-2em}
    \label{fig:complexity_crop}
\end{figure}

The results of this experiment are shown in Table \ref{tab:complexity}.
We observe that our method obtains the lower complexity compared to VecRoad \cite{Tan2020}
and DeepRoadMapper \cite{Mattyus2017}. This is mainly due to
our sparse approach to the regression of junctions, as shown by the average number of nodes.
In Figure \ref{fig:complexity_crop}, we show the way our method
finds an optimal representation of a neighborhood compared to the very high number of nodes
and edges found by VecRoad \cite{Tan2020}.

\begin{table}[h!]
    \centering
    \small
    \setlength\tabcolsep{4pt}
    \begin{tabular}{c|c|c|c|c|c}
    Method & Nodes & Edges & Total & APLS & Complexity \\ \hline
    VecRoad \cite{Tan2020} & 29620 & 61042  & 90662 & \textbf{64.59} & 1404\\
    DRM \cite{Mattyus2017} & 6071  & 11963  & 18034 & 21.27 & 848\\
    RoadTracer \cite{Bastani2018} & 8263 & 17044 & 25306 & 45.09 & 561 \\ \hline
    Ours & 4343 & 17273 & 21615 & 46.93 & \textbf{461}\\
    \end{tabular}
    \caption{Comparison of average graph complexity scores (Total elements divided by APLS. Lower is better). Nodes (resp. Edges) is the average number of nodes (resp. edges) in the
    output graphs over the whole RoadTracer test set.
    DRM = DeepRoadMapper. Our method achieves the lowest complexity compared
    to other approaches, which make it a good compromise between accuracy and
    compactness of graphs.}
    \vspace{-2em}
    \label{tab:complexity}
\end{table}

\section{Limitations}

\begin{figure}[h!]
    \centering
    \includegraphics[height=0.4\columnwidth]{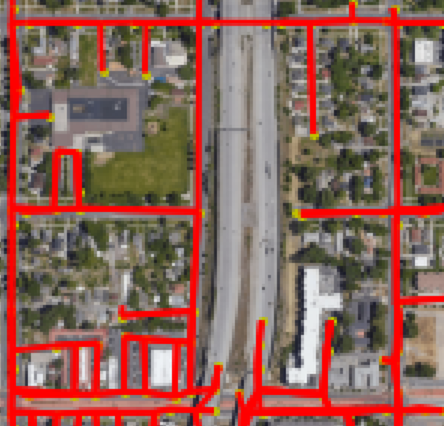}
    \includegraphics[height=0.4\columnwidth]{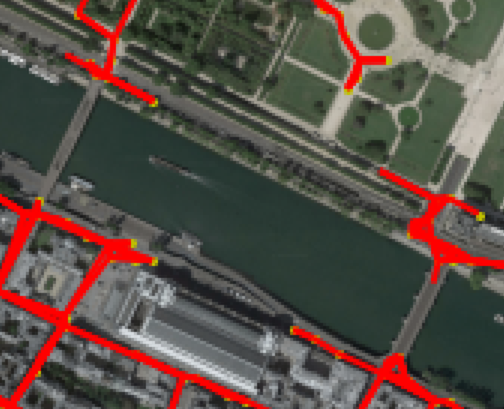}
    \caption{Examples of long sections of road (e.g. highways, bridges) without points 
    of interest, which are occasionally missed by our method.}
    \label{fig:limitations}
    \vspace{-1em}
\end{figure}

Our method comes with some limitations. The first one,
as said in previous sections, is the difficulty to 
work in very dense areas with lots of intersections.
Since multiple intersections might fall into the same
detection cell, our network will only be able to detect
one of them, which can lead to shifted junctions.
The second limitation is the inability to be trained on
mask-based datasets such as DeepGlobe \cite{demir2018deepglobe}.
Our network can only be trained on graphs, and such data
might not always be available.
Finally, some long sections
of road such as highways or bridges might lack points of interest
since they do not have any junctions. Our network
can miss these sections of road in the final graph as shown in Figure \ref{fig:limitations}
and in some parts of Figure \ref{fig:teaser}.

\section{Conclusion}

We proposed a novel architecture for single-shot extraction of road graphs from satellite images. Our method first predicts sparse interest points using an image CNN feature extractor as well as a junction detection head, and an offset regression head, trained jointly to identify candidate road intersections in a lower-resolution image and predict their position in the high-resolution image. We then combine the extracted image features and regressed point coordinates with a graph neural network to combine local and global information and infer the unknown graph structure. Our method combines aspects of graph learning and link prediction to score candidate connections between road junctions and infer the road graph. We demonstrate competitive performance with state of the art methods on three reference metrics while achieving significantly lower graph complexity and inference times (up to 91x faster than iterative models) compared to iterative and segmentation-based methods. 

Even though the experimental validation of our approach was focused on road networks,
our single-shot graph extraction framework can be applied to other types of problems.
An example would be the extraction of blood vessels in medical images \cite{fraz2012blood}.
An interesting follow-up to road extraction
could be to not only decide if edges exist or not in the final graph,
but also infer edge attributes like road types, amount of traffic, 
or speed limits, as done in \cite{gharaee2021graph}.
We also believe that our method could be adapted to tackle map update problems 
that have been presented in newly released datasets \cite{bastani2021beyond}.
In addition, our method can further be combined with existing model compression techniques, both from the Euclidean and Geometric deep learning literature, to pave the way for on board single-shot extraction of road graphs on edge devices.

\section*{Acknowledgements}

G.B. is funded
by the CIAR project at IRT Saint Exupéry.
M. B. was supported by a Department
of Computing scholarship from Imperial College London,
and a Qualcomm Innovation Fellowship.
The authors are grateful to the OPAL infrastructure from 
Université Côte d'Azur for providing resources and support.
This work was granted access to 
the HPC resources of IDRIS under the allocation 2020-
AD011011311R2 made by GENCI.

\appendix

\section{Impact of choosing a smaller backbone}

Feature extractors (backbones) come in different sizes to suit
different kinds of performance targets. For example, some applications
like edge computing might accept a reduction on accuracy in favor of faster inference time and higher throughput.
In order to evaluate the impact of choosing a smaller backbone on our method,
we take the network presented in the main text and replace the ResNet-50 backbone
with ResNet-18. 

\begin{table}[h!]
    \centering
    \begin{tabular}{c|c|c}
    Method & APLS & Single image time (s) \\ \hline
    Ours (R18) & 38.80 & \textbf{19.6} \\
    Ours (R50) & \textbf{40.45} &  20.9 \\
    \end{tabular}
    \caption{Run-time in seconds on the RoadTracer Dataset.
    The single image score is averaged over the whole test set. 
    Using a smaller backbone (R18) is slightly faster but causes a small drop in APLS.}
    \label{tab:benchmark_r18}
\end{table}

The results of this experiment are presented in Table \ref{tab:benchmark_r18}.
We observe that using a ResNet-18 backbone is slightly faster but results in a slightly lower
APLS score. This means that our method works with smaller backbones and thus can be 
used on a variety of low power devices, especially ones that have a low amount of memory, such
as the Nvidia Jetson family of devices.

\section{Impact of edge detection threshold}

\begin{figure}[h!]
    \centering
    \includegraphics[width=\columnwidth]{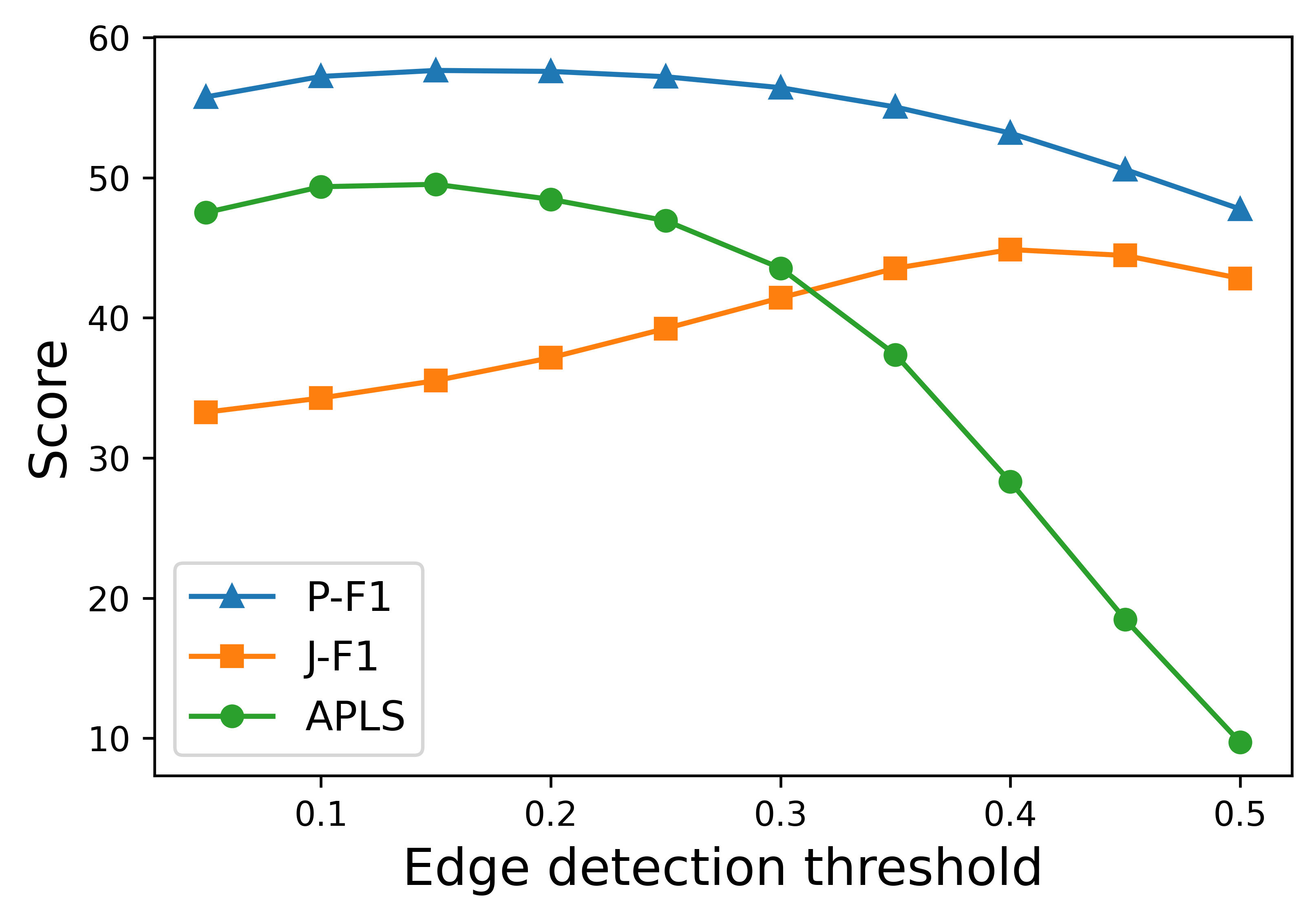}
    \caption{J-F1, P-F1 and APLS scores for our main model, at a range of edge detection thresholds.}
    \label{fig:threshold}
\end{figure}

In this section, we study the impact of the edge detection threshold, which can
be freely chosen between 0 and 1. To do this, we compute the three main accuracy metrics
for a range of thresholds varying between 0 and 0.5, using our main model.
Setting a high detection threshold favors the Precision of edges, while
setting a low threshold favors the Recall.
Figure \ref{fig:threshold} shows the results of this experiment. 
We observe that the P-F1 (pixel metric) and APLS are better when detecting more edges, while
the J-F1 (junction metric) is slightly better when detecting fewer edges. This makes 
sense as the J-F1 is based on having the correct number of edges for each junction,
while the APLS should benefit from more path options. The P-F1 may be higher at low thresholds
simply because we find more road pixels.
This experiment shows that finding and setting an appropriate edge threshold for each model is important
in order to obtain the best accuracy and the best compromise between the three metrics.

\section{On the J-F1 metric}

As previously said in the Method and Limitations sections of our main text,
our method is only able to find a single point-of-interest per output cell,
which can lead to merged junctions in the output graph. 
In addition, as seen in the Experiments section of the main text,
our method will inherently find a lower number of junctions since it
is tailored towards the extraction of a sparse graph.
Since the J-F1 score is based on a matching of junctions within a certain radius,  
we believe that these design choices significantly affect this metric,
which leads to our method scoring lower than some other approaches.

\newcommand{\cmark}{\ding{51}}%
\newcommand{\xmark}{\ding{55}}%

\begin{table*}[h!]
    \centering
    \setlength\tabcolsep{4pt}
    \begin{tabular}{c|c|c|c|c|c|c|c|c|c|c|c}
    \# & Pre-trained & Init Graph & GNN & Classifier & Raw feat & Wait & Train $j_{thr}$ & J-F1 & P-F1 & APLS & Sum \\ \hline
    
1 & \cmark & Baseline & \xmark & Bilinear   & & & & \textbf{40.33} & 55.93 & 43.95 & 140.21\\
2 & & Baseline & \xmark & Bilinear   & & & & 36.03 & 53.87 & 41.64 &  131.54 \\
3 & & Baseline & \xmark & Bilin small  & & & & 32.24 & 50.4 & 36.11 & 118.75\\ \hline

4 & \cmark & Baseline & \xmark & MLP   & & & & 36.92 & 53.43 & 41.17 & 131.53\\ 
5 &  & Baseline & \xmark & MLP   & & & & 35.46 & 56.03 & 45.35 &  136.84\\ \hline

6 & \cmark & Complete &  EdgeConv & Bilinear   &  & & & 37.69 & 56.04 & 45.39 & 139.13\\
7* & \cmark & Complete &  EdgeConv & MLP   & & & & 39.23 & 57.2 & 46.93 & 143.36\\ \hline

8 & \cmark & Dynamic &  EdgeConv & Bilinear   &  &  &  & 35.63 & 56.29 & 46.04 &  137.96 \\
9 & \cmark & Dynamic &  EdgeConv & Bilinear   & &  &  0.3 & 37.90 & 55.5 & 43.68 & 137.08\\  
10 & \cmark & Dynamic &  EdgeConv & Bilinear   & \cmark &  &  & 37.32 & 52.07 & 38.51 &  127.91 \\
11 & \cmark & Dynamic &  EdgeConv & Bilinear   & \cmark &  &  0.3 & 37.92 & 52.37 & 40.27 & 130.55\\ \hline

12 & \cmark & Dynamic &  EdgeConv & MLP & & & & 38.44 & 56.7 & 46.21 &  141.35 \\ 
13 & \cmark & Dynamic &  EdgeConv & MLP & \cmark & & & 34.95 & 53.75 & 41.57 &  130.27 \\
14 & \cmark & Dynamic &  EdgeConv & MLP   & \cmark &  &  0.3 & 35.54 & 56.5 & 45.91 & 137.95\\ \hline

15 & \cmark & kNN-4  &  EdgeConv & Bilinear   & & & & 37.30 & 54.25 & 42.28 & 133.83\\
16 & \cmark & kNN-4  &  EdgeConv & MLP   & & & & 37.41 & \textbf{57.54} & \textbf{48.71} & \textbf{143.66}\\ \hline

17 & \cmark & Complete  &  DeepGCN & MLP   & \cmark & & & 37.17 & 55.43 & 42.59 & 135.19 \\
18 & \cmark & Complete  &  1 Conv + DeepGCN & MLP   & & & & 38.35 & 57.11 & 47.12 & 142.58 \\
19 & \cmark & Complete  &  4 Conv + DeepGCN & MLP  & & & & 37.89 & \textbf{57.71} & \textbf{48.84} & \textbf{144.44}\\ \hline

20 & & Dynamic &  EdgeConv & Bilinear   & &  &  & 34.52 & 50.74 & 35.52 &  120.78 \\
21 & & Dynamic &  EdgeConv & Bilinear   & \cmark &   30 & & 33.23 & 46.26 & 30.71 & 110.20\\ 
22 & & Dynamic &  EdgeConv & Bilinear   &  &  30 &  & 32.03 & 50.94 & 37.31 &  120.28 \\
23 & & Dynamic &  EdgeConv & Bilinear   & \cmark &   30 &   0.3 & 36.89 & 49.02 & 32.13 & 118.04\\ \hline

24 & & Dynamic &  EdgeConv & MLP & &  &  & 36.35 & 56.31 & 45.55 & 138.21 \\
25 & & Dynamic &  EdgeConv & MLP & & 30  &  & 36.54 & 55.72 & 43.97 &  136.22 \\
26 & & Dynamic &  EdgeConv & MLP   & &  30 &   0.3 & 34.32 & 55.41 & 45.65 & 135.38\\
27 & & Dynamic &  EdgeConv & MLP   & &  10 &   0.3 & 35.36 & 54.96 & 44.44 & 134.76\\
28 & & Dynamic &  EdgeConv & MLP   & \cmark &   30 &   0.3 & 34.48 & 55.02 & 42.15 & 131.65\\  
29 & & Dynamic &  EdgeConv & MLP   & \cmark &   30 & & 35.73 & 54.9 & 41.91 & 132.54\\ 
30 & & Dynamic &  EdgeConv & MLP   & \cmark &   10 &   0.3 & 34.90 & 56.03 & 44.60 & 135.54\\

    \end{tabular}
    \caption{Variations on our model.
    Pre-trained = use of pre-trained ResNet and junctionness/offset branches for faster training.
    Raw feat = GNN applied directly to ResNet features (no node feature convolutions).
    Wait = Number of epochs where only the junction-ness and offset branches are trained before training the edge branch (default: 0).
    Train $j_{thr}$ = junction-ness threshold used during training (default: 0.5).
    Sum = J-F1 + P-F1 + APLS.
    *variation presented in the main text}
    \label{tab:gnn}
\end{table*}

\section{GNN model choices and ablation study}

In this section, we perform a comparative study of different design decisions for the GNN portion of the model. 

We compare the following choices for the construction of the road graph:
\begin{enumerate}
    \item Using the complete graph as the supporting graph for each GNN layer and treating the problem as a pure edge classification task
    \item Initializing the road graph by $k$-NN search (k = 4) on the output of the node feature branch (dim = 256) to which we concatenate the Cartesian coordinates of the junctions (dim = 2)
    \item Dynamic construction of the graph at each layer by $k$-NN search (k = 4) on the layer's input (\ie, as described in point 2 for the first layer, and on the output features of the previous layer for subsequent layers)
\end{enumerate}

We also compare two choices of edge scoring function:
\begin{enumerate}
    \item A 2-layer MLP ($\text{FC}(256) \rightarrow \text{FC}(256)$ where FC($N$) denotes a fully-connected layer with $N$ output features)
    \item A $256 \times 256 \times 1$ bilinear layer
\end{enumerate}

We use a three-layer graph neural network based on the EdgeConv operator with BatchNormalization and ReLU activations. We parameterize the EdgeConv operators each with a linear layer with 256 output features.  We also report the results of baseline models trained without graph convolutions.

Finally, we compare the 3-layer EdgeConv networks with DeepGCNs \cite{li2019deepgcns,Li2020Deeper} using the Generalized Graph Convolution (GENConv) operator \cite{Li2020Deeper}, layer normalization \cite{ba2016layer} and ReLU activations. We apply the DeepGCN layers sequentially on:
\begin{enumerate}
    \item The raw image features produced by the backbone (DeepGCN)
    \item The output of one convolutional layer in the node feature branch (1 Conv + DeepGCN)
    \item The output of 4 convolutional layers in the node feature branch (4 Conv + DeepGCN) as for the EdgeConv-based models
\end{enumerate}

In case (1) we used 7 layers of (GENConv $\rightarrow$ LayerNorm $\rightarrow$ ReLU), in case (2) we used 6 layers, and in case (3) we used 3 layers, so as to keep model capacity comparable with the EdgeConv models applied on the full node feature branch. We applied the DeepGCNs on the complete graphs only. 

For each of these experiments, we report the J-F1, P-F1 and APLS metrics at the edge threshold that maximizes their sum. Table \ref{tab:gnn} shows the results of these experiments. To enable faster experimentation, we initialized some of the models using the pre-trained weights of a "Baseline MLP" model trained on the complete graph, such models are denoted by a checkmark in the "Pre-trained" columns of Table \ref{tab:gnn}.

We can draw the following conclusions from the experimental results shown:
\begin{itemize}
    \item The baseline with the bilinear classifier outperforms the baseline with a 2-layer MLP, which is expected since it has a much larger number of parameters. While the improvement is noticeable, the model loses compactness and efficiency.
    \item All three constructions of the graph (complete, static $k$-NN and fully dynamic) are able to perform well and to outperform the MLP and the Bilinear baseline. Notably, the models that combine a GNN with an MLP classifier outperformed the ones with the same GNNs but a bilinear classifier (scoring function). The entire GNN adds fewer parameter than changing the MLP for a bilinear layer, and yet can bring larger performance deltas, which further motivates our choice of using an MLP scoring function.
    \item The best performing GNN with the EdgeConv operator used a fixed $4$-NN graph support for message passing (but still scored all possible edges for the final graph). However, choosing the right value of $k$ adds another element to hyperparameter tuning of the model and comes at the disadvantage of reduced robustness to cases where images have no junctions to detect (\eg satellite images of large bodies of water such as lakes). We therefore chose to report the performance of the simpler model in the main text, while showcasing that sparse graph priors may indeed lead to better road graph reconstructions.
    \item Getting rid of the node features branch reduces the number of trainable parameters but appears to be detrimental to the performance in most cases. Experiments with deeper GNNs applied directly on the output of the backbone showed increased performance compared to shallower GNNs, although the graph construction differs (l. 14, 17)%
    \item Compared to 3-layer EdgeConv networks on the complete graphs, 3-layer DeepGCNs applied following either a 1-layer or 4-layer node feature branch performed better. These models also outperformed the dynamic graph models and the 3-layer EdgeConv applied on a sparse $4$-NN graph (by a slimmer margin in the latter case). The DeepGCNs applied on raw features outperformed some of the shallower models trained with a node feature branch, but did not match the best performing models.
    \item All DeepGCNs outperformed the MLP baseline, which indicates that, 
    keeping the edge scoring function the same, replacing all (DeepGCN, l. 17)%
    or most (1 Conv + DeepGCN, l. 18)%
    2D convolutions with graph convolutions leads to increased performance compared to a model that only uses 2D convolutions.
\end{itemize}

\paragraph{Deep Graph Convolutional Networks} We suspect the last two points are due to two effects: first, 2D convolutions can be seen as special cases of graph convolutions applied to the 2D lattice graph while enforcing translation equivariance; their inductive bias is well suited to the processing of images. In contrast, we applied the GNNs (EdgeConv or DeepGCN) on the (dynamic or complete) graphs of detected junctions, which means the GNNs do not have access to the neighboring pixel's context whereas the 2D Euclidean convolutions do. We believe this contributes to the reduced performance of all graph-convolutional models compared to models that combine 2D convolutions and graph convolutions. Additionally, the higher performance of the graph convolution models compared to only using 2D convolutions - especially on the APLS metric - show the contribution of the graph-based models.

Regarding the relative performance of DeepGCNs compared to shallower EdgeConv models, the graphs extracted from the images are small: they have at most $16 \times 16 = 256$ nodes and $(256)*(256-1)/2 = 32640$ edges. The increase in receptive field that comes with deeper models will lose effectiveness once the entire graph is covered at a given layer. Furthermore, the experiments we report with DeepGCNs in Table \ref{tab:gnn} are done using the complete graph as support. Messages from each node reach all other nodes in a single iteration, which reduces the benefits one can glean from using deeper models, and makes the GNNs more susceptible to the smoothing problem \cite{Li2018}. We therefore decided to report results using shallow GNNs using the EdgeConv operator which is well suited to learning edge features and to learning on dynamic graphs. Further work will investigate using DeepGCNs on sparse graphs as well as on graphs built on larger images.

\paragraph{Junctionness threshold} In addition to the ablation study, we evaluated the impact of the junction-ness threshold $j_{thr}$ used during training. 
Lowering this threshold seems to have a positive impact on the three metrics. Our intuition is that when the junction-ness threshold is lower,
the subsequent GNN is presented with more nodes and a larger number of possible edges for each image, and is thus trained better and faster.

\section{Other uses of our method}

\begin{figure}[h!]
    \centering
    \includegraphics[width=0.49\columnwidth]{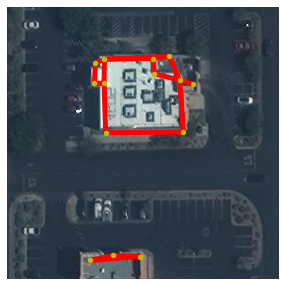}
    \includegraphics[width=0.49\columnwidth]{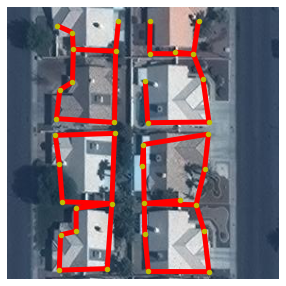}
    \caption{Our method can be used for other tasks, such as building extraction. 
    This example is taken from the CrowdAI Mapping Challenge dataset \cite{mohanty2020deep}}
    \label{fig:building}
\end{figure}

Our method is generic in the sense that it could be used for 
applications other than road extraction. In fact, it could be suitable for
any image to 2D graph application. For example, our method could be used for
blood vessel extraction as done in \cite{Ventura2018}.
Another example is the polygonal extraction of buildings in aerial 
images. Just like for road extraction, existing 
methods either rely on an iterative process \cite{Li2019} or on 
post-processing of a pixel-based segmentation \cite{Li_2020_CVPR}.
Thus, our method is able to provide the same advantages in this task
as for road extraction.
Figure \ref{fig:building} shows an example of building extraction using
our method.

\section{Qualitative results}

\begin{figure*}[p!]
    \centering
    \includegraphics[trim={2mm 2mm 2mm 2mm},width=0.32\textwidth]{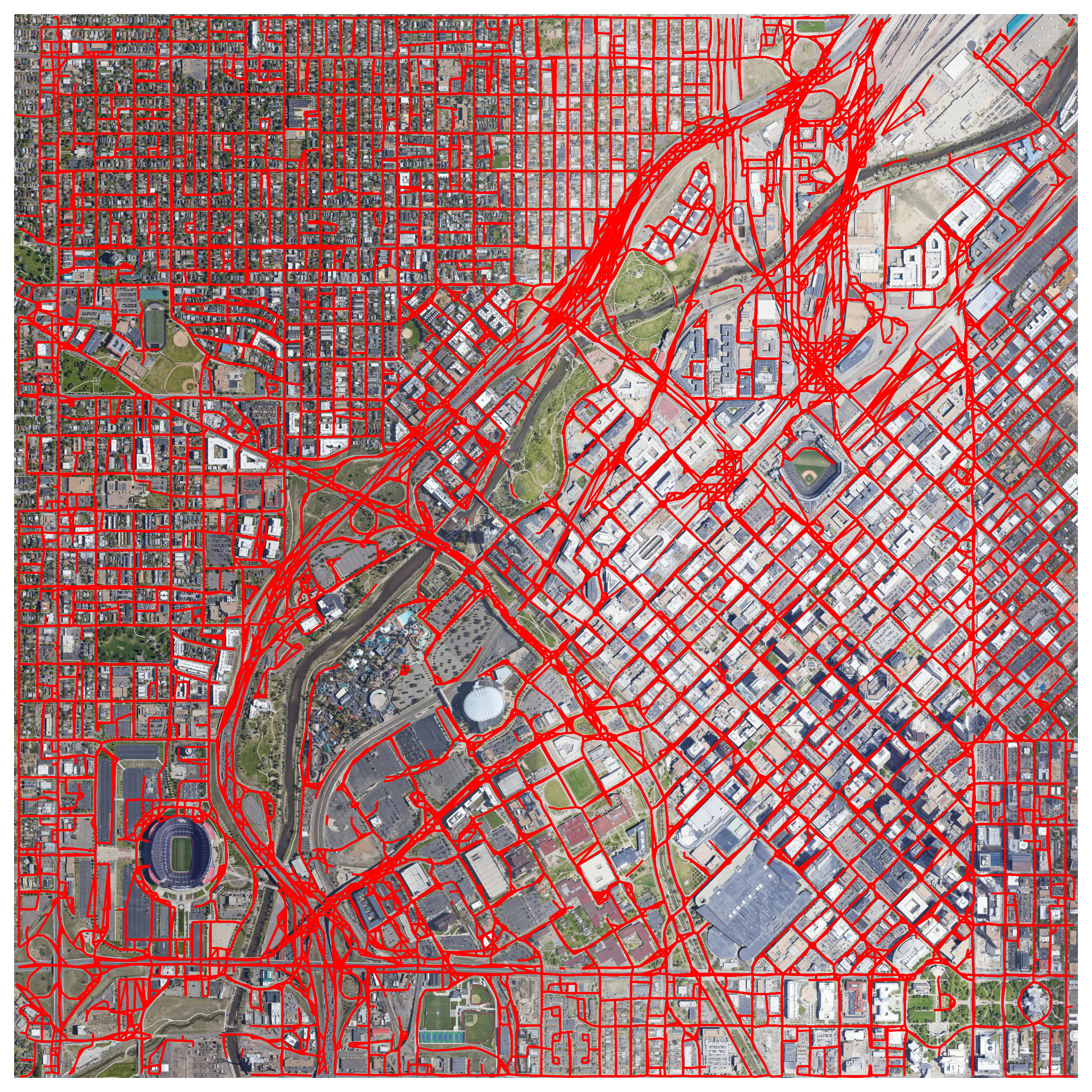}
    \includegraphics[width=0.32\textwidth]{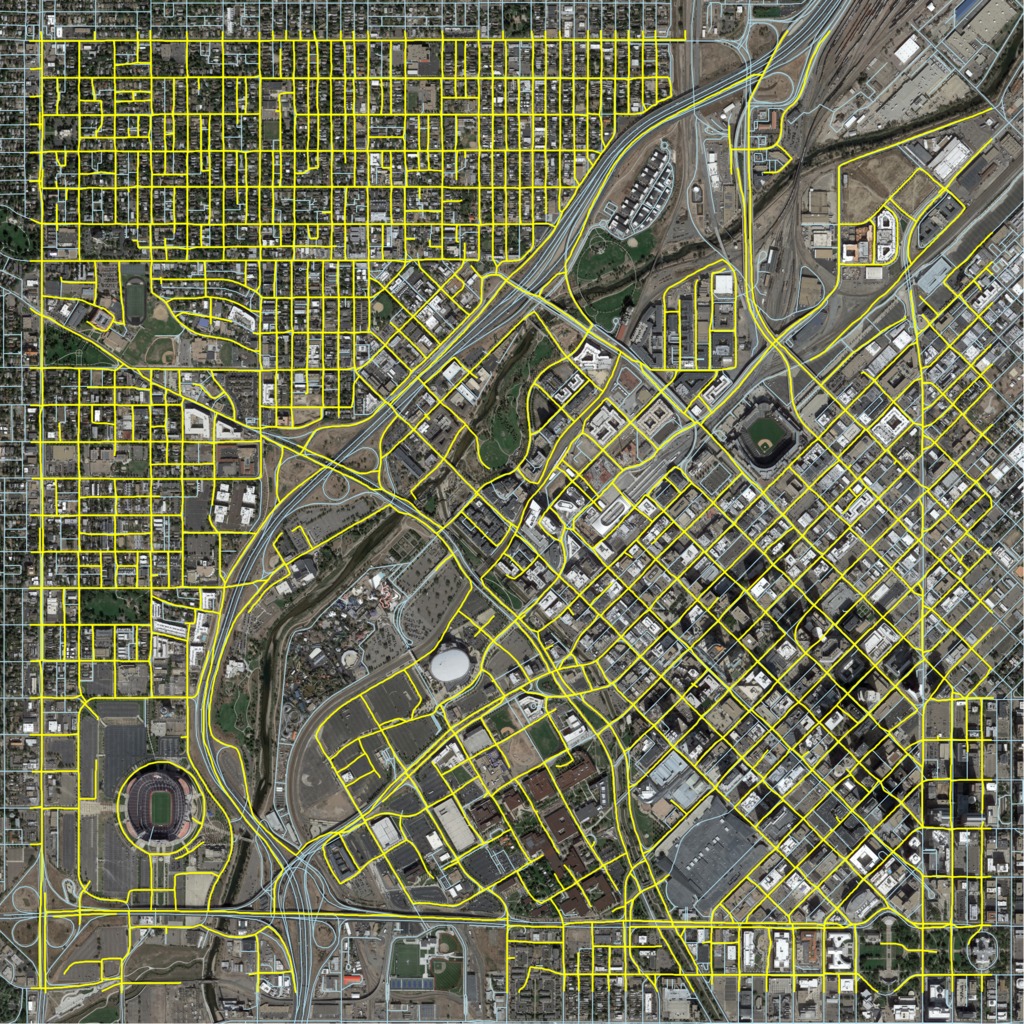}
    \includegraphics[width=0.32\textwidth]{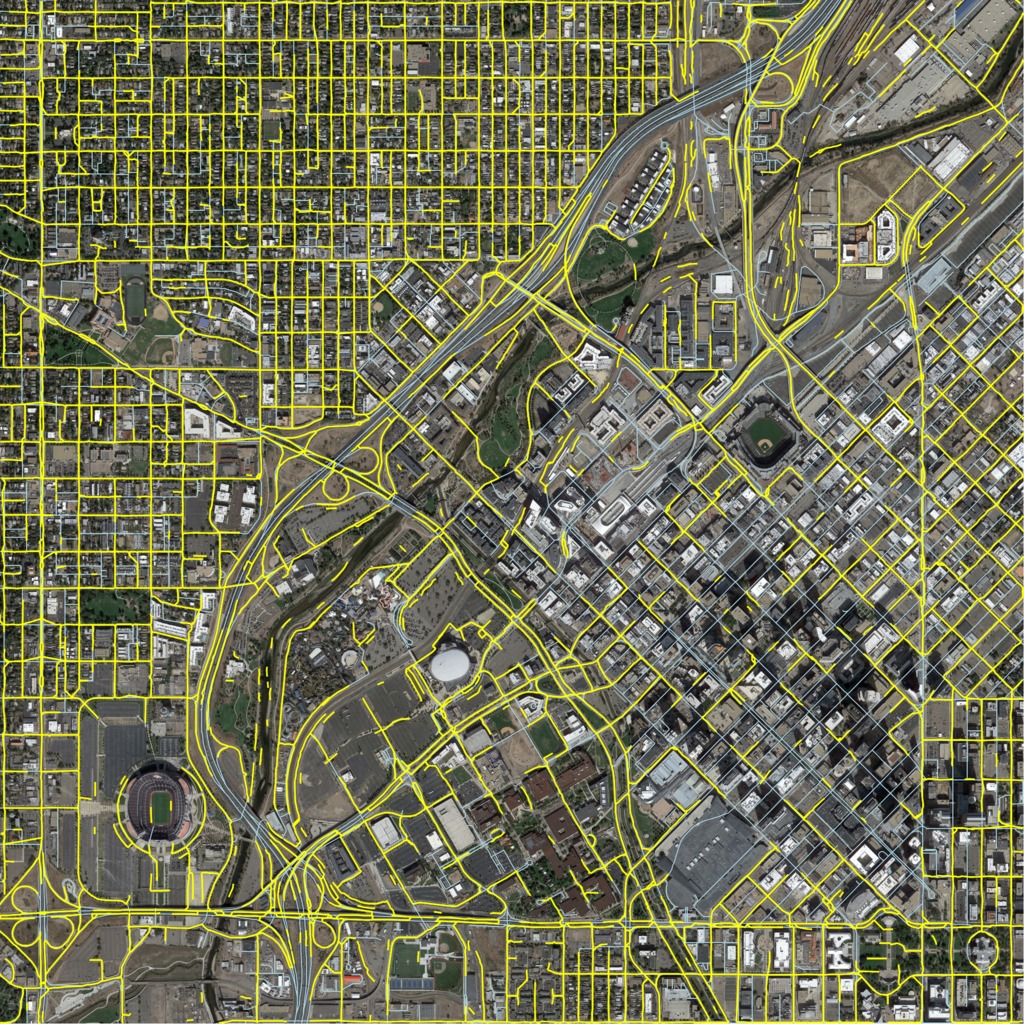}\hfill \\
    
    \includegraphics[trim={2mm 2mm 2mm 2mm},width=0.32\textwidth]{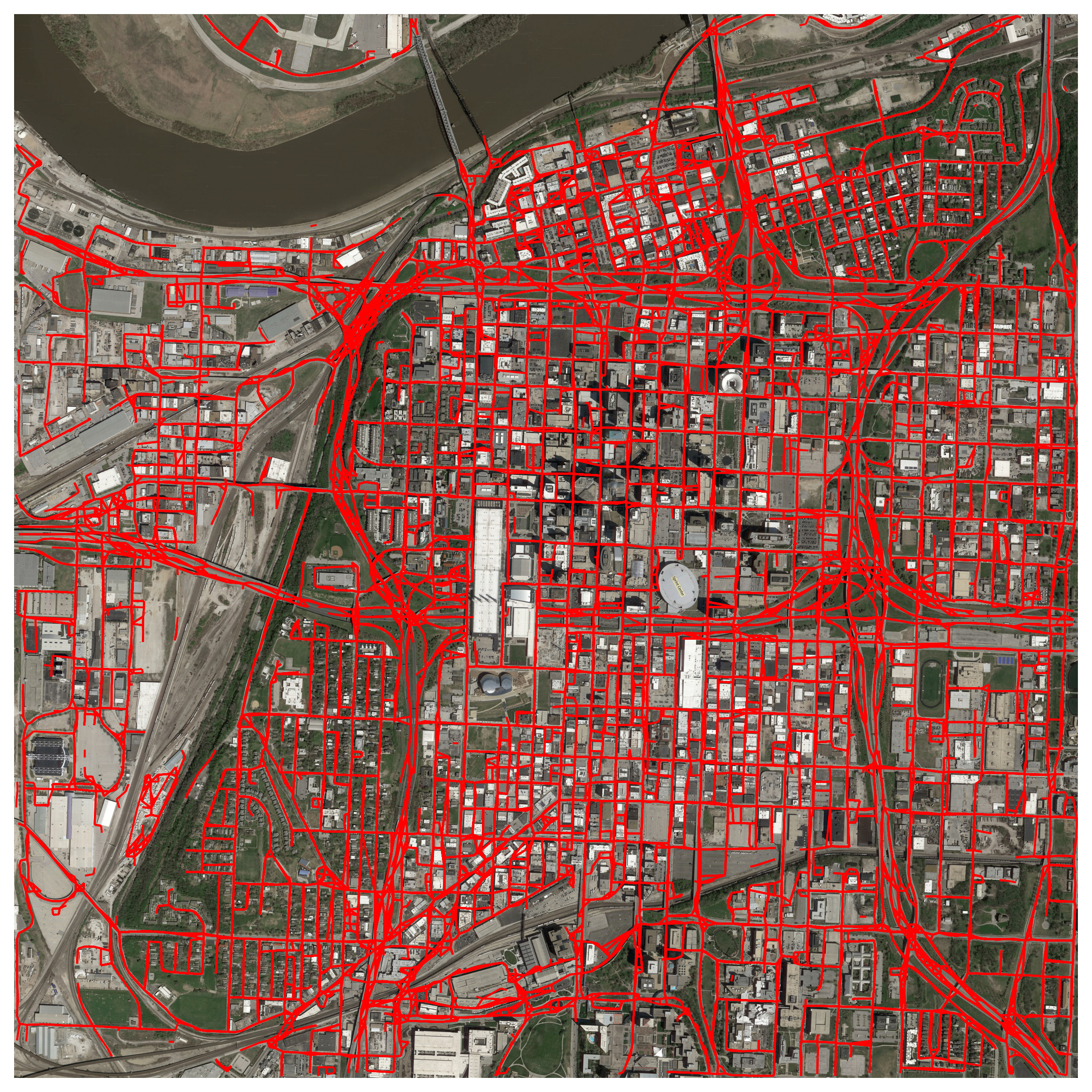}
    \includegraphics[width=0.32\textwidth]{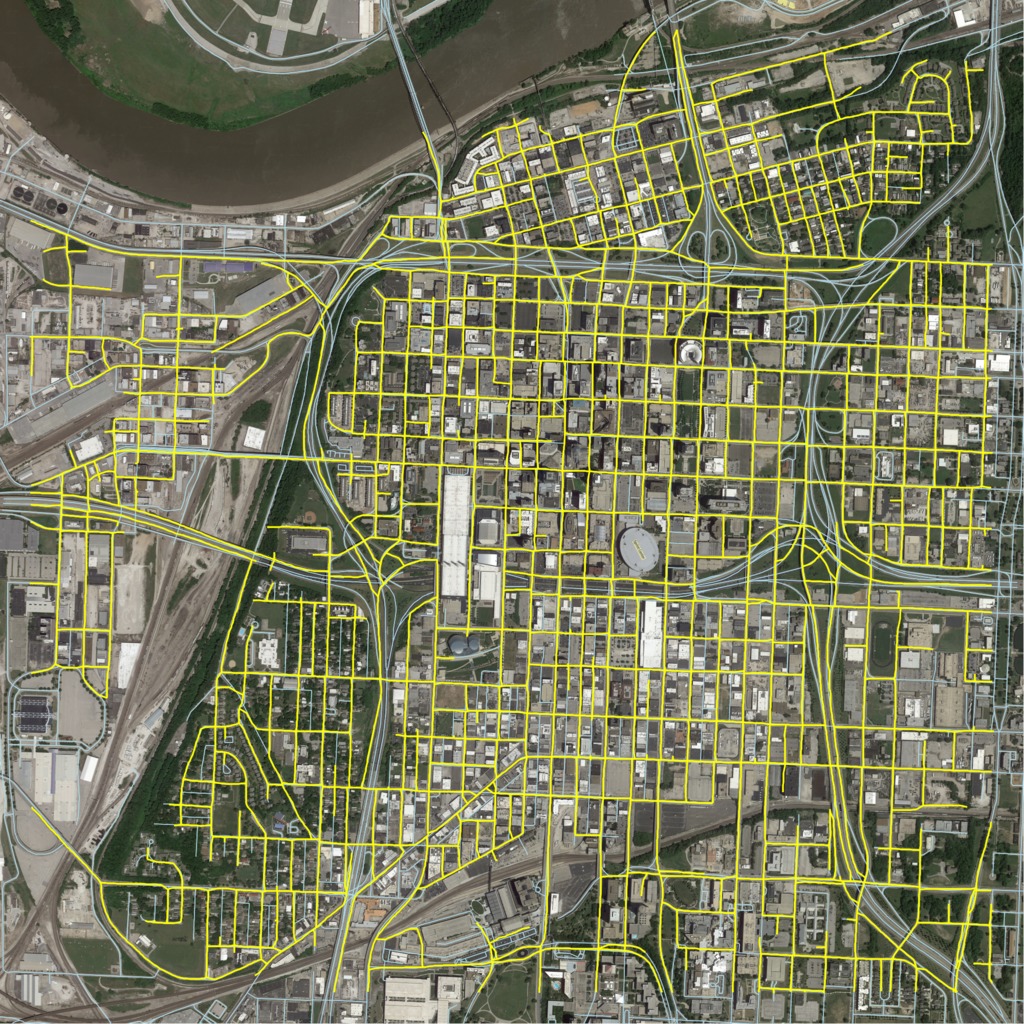}
    \includegraphics[width=0.32\textwidth]{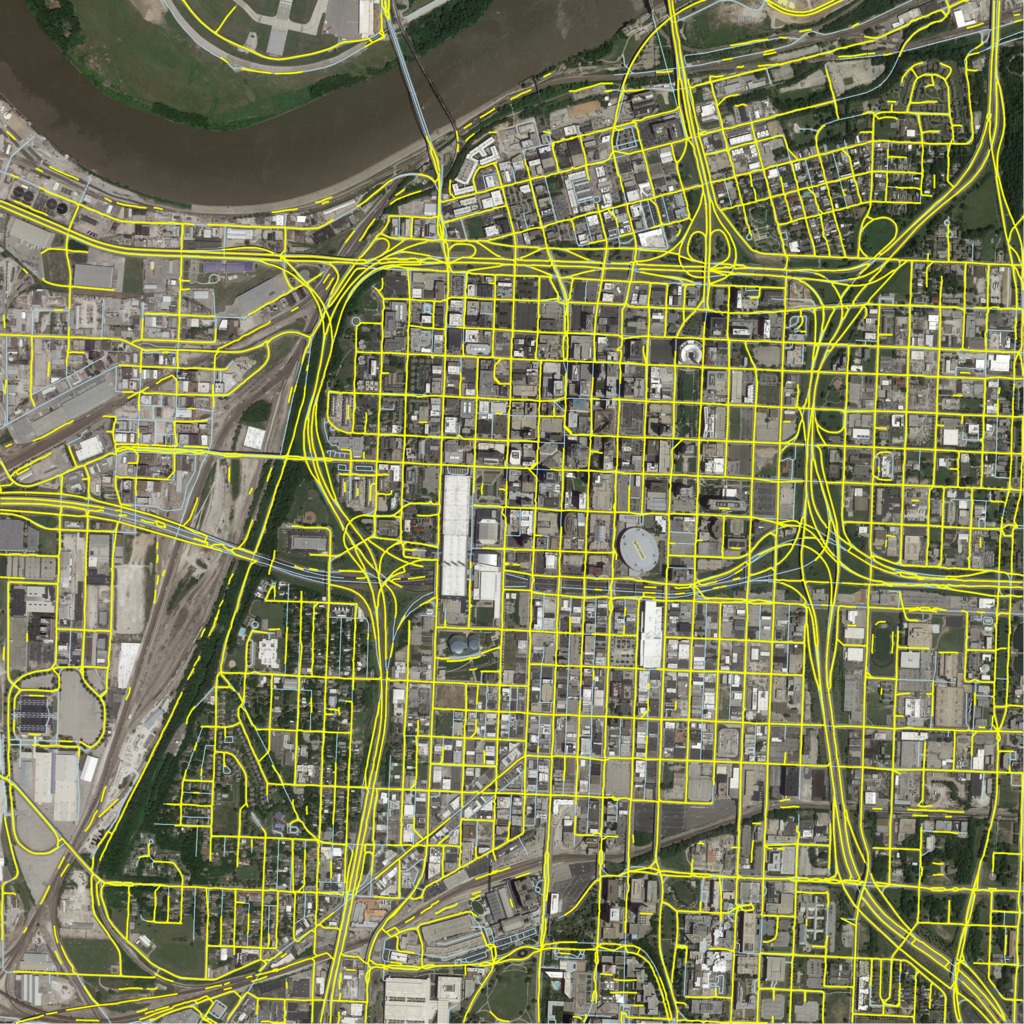}\hfill \\
    
    \includegraphics[trim={2mm 2mm 2mm 2mm},width=0.32\textwidth]{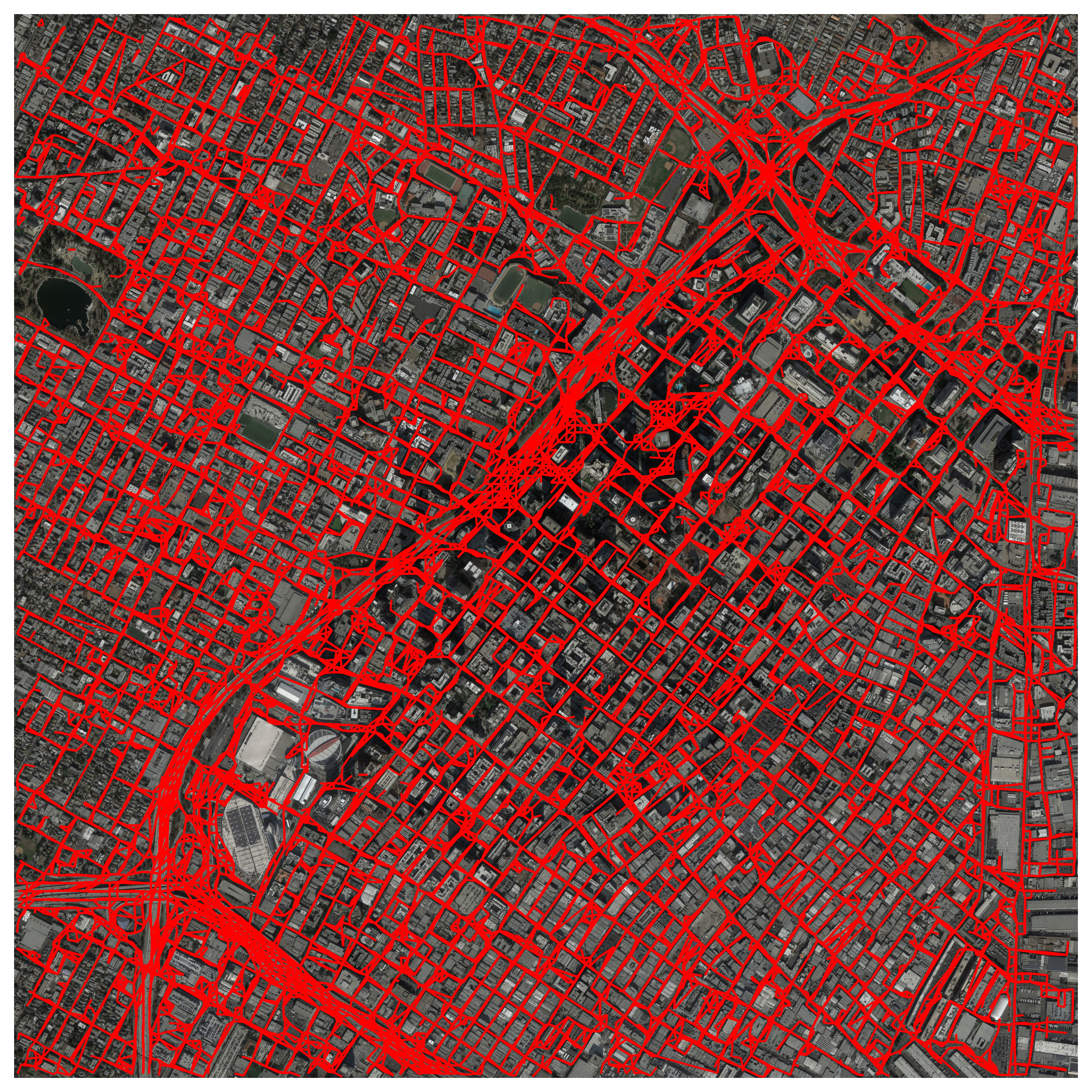}
    \includegraphics[width=0.32\textwidth]{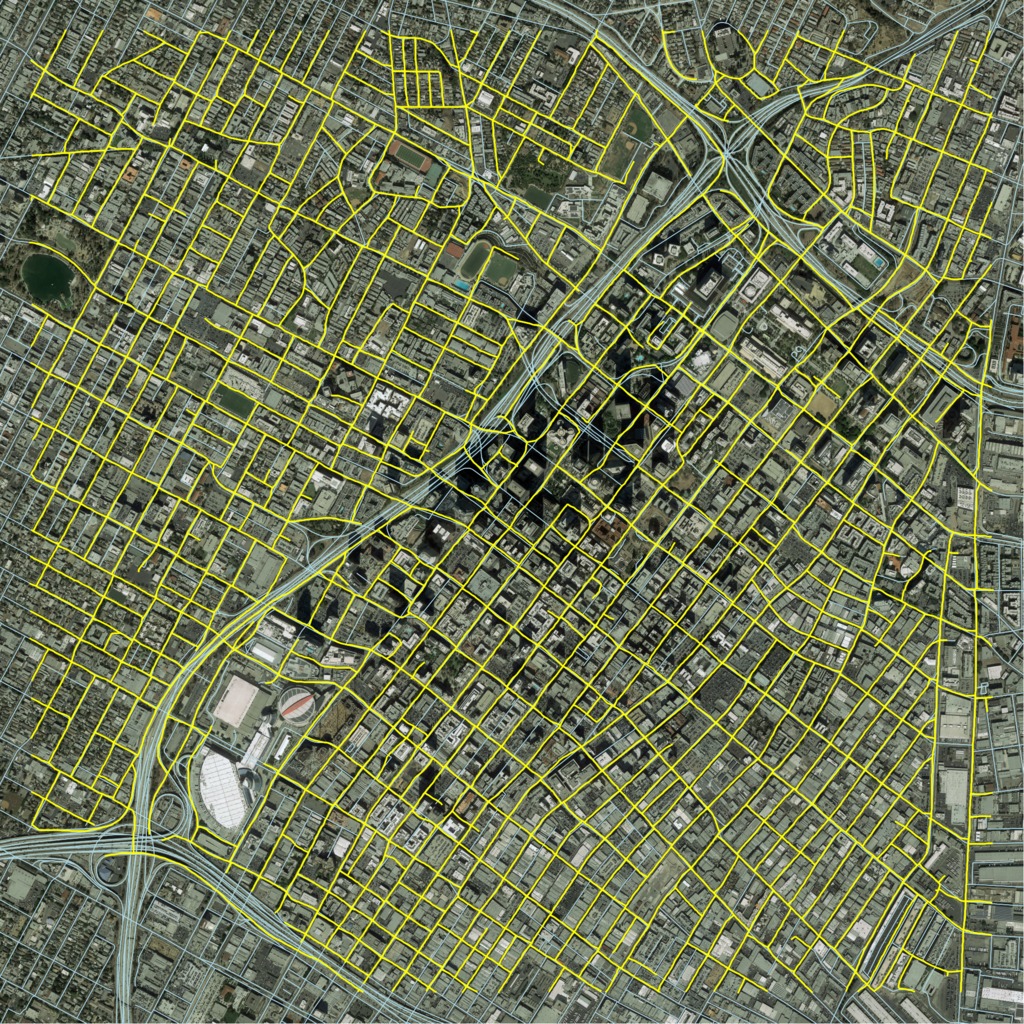}
    \includegraphics[width=0.32\textwidth]{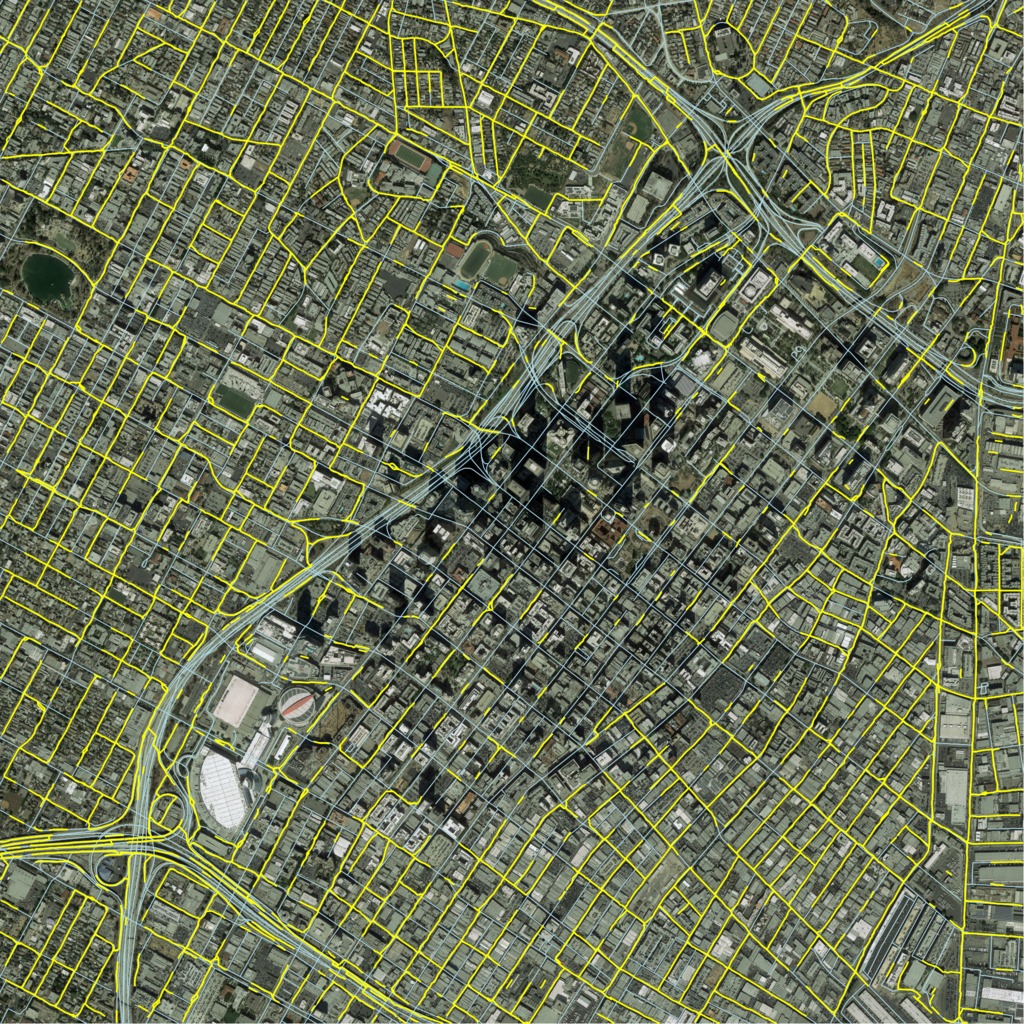}\hfill \\
    
    \includegraphics[trim={2mm 2mm 2mm 2mm},width=0.32\textwidth]{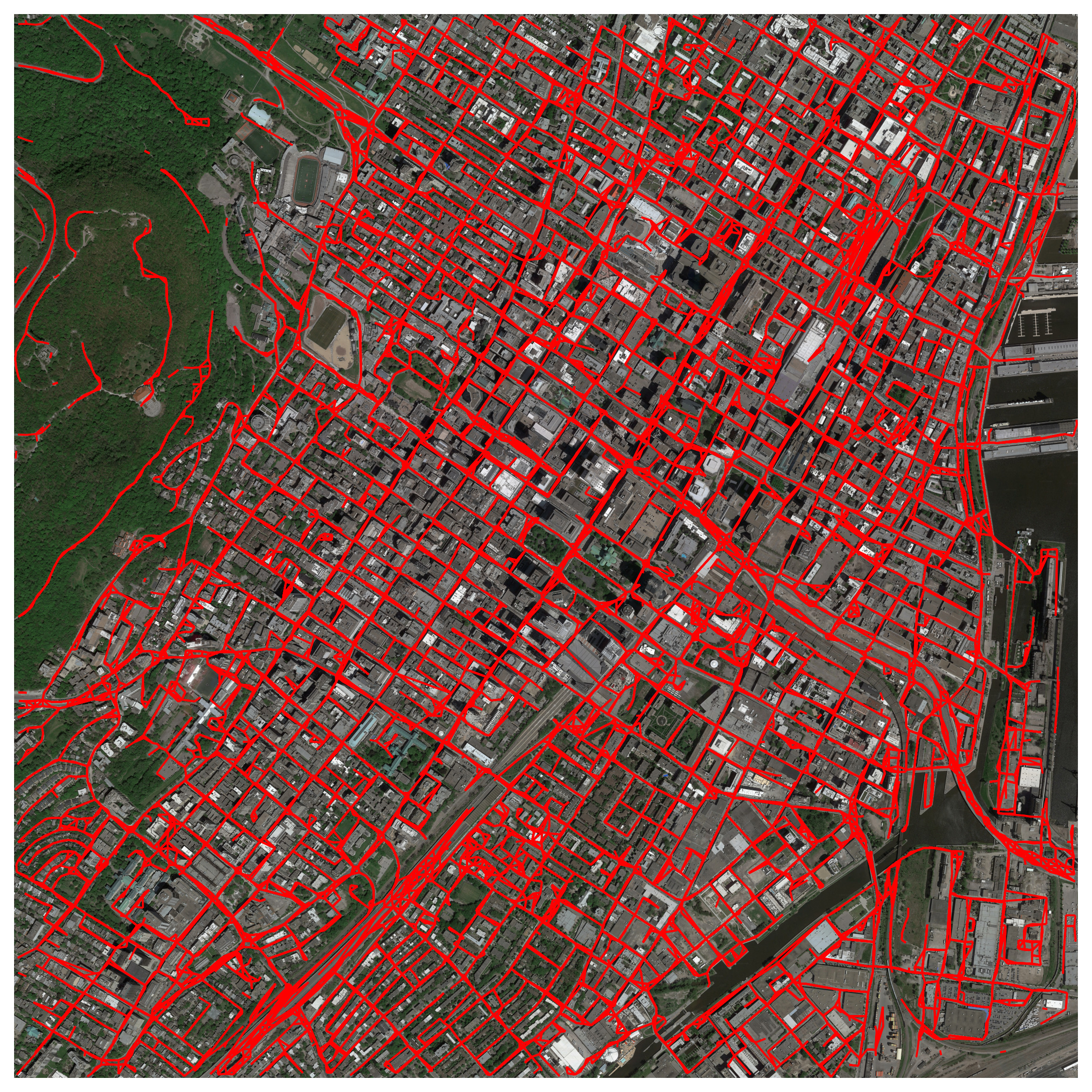}
    \includegraphics[width=0.32\textwidth]{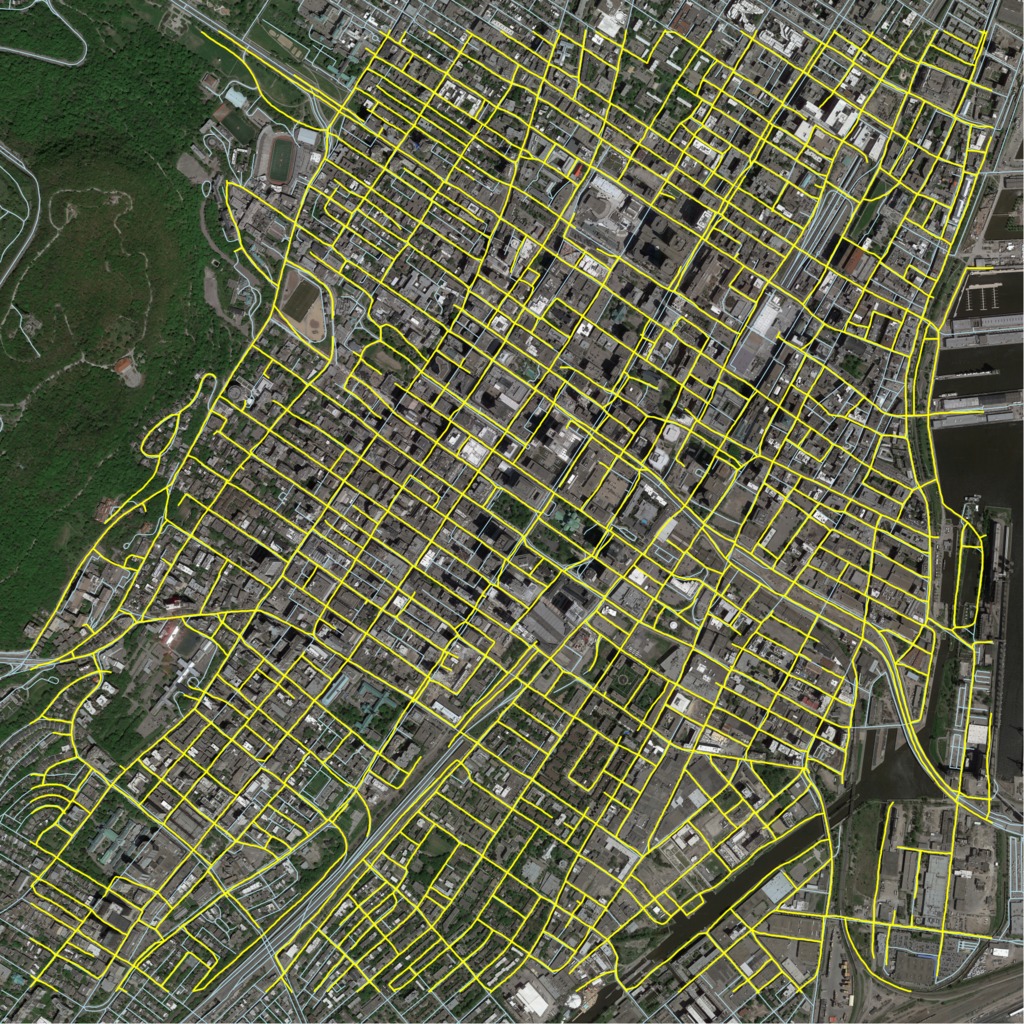}
    \includegraphics[width=0.32\textwidth]{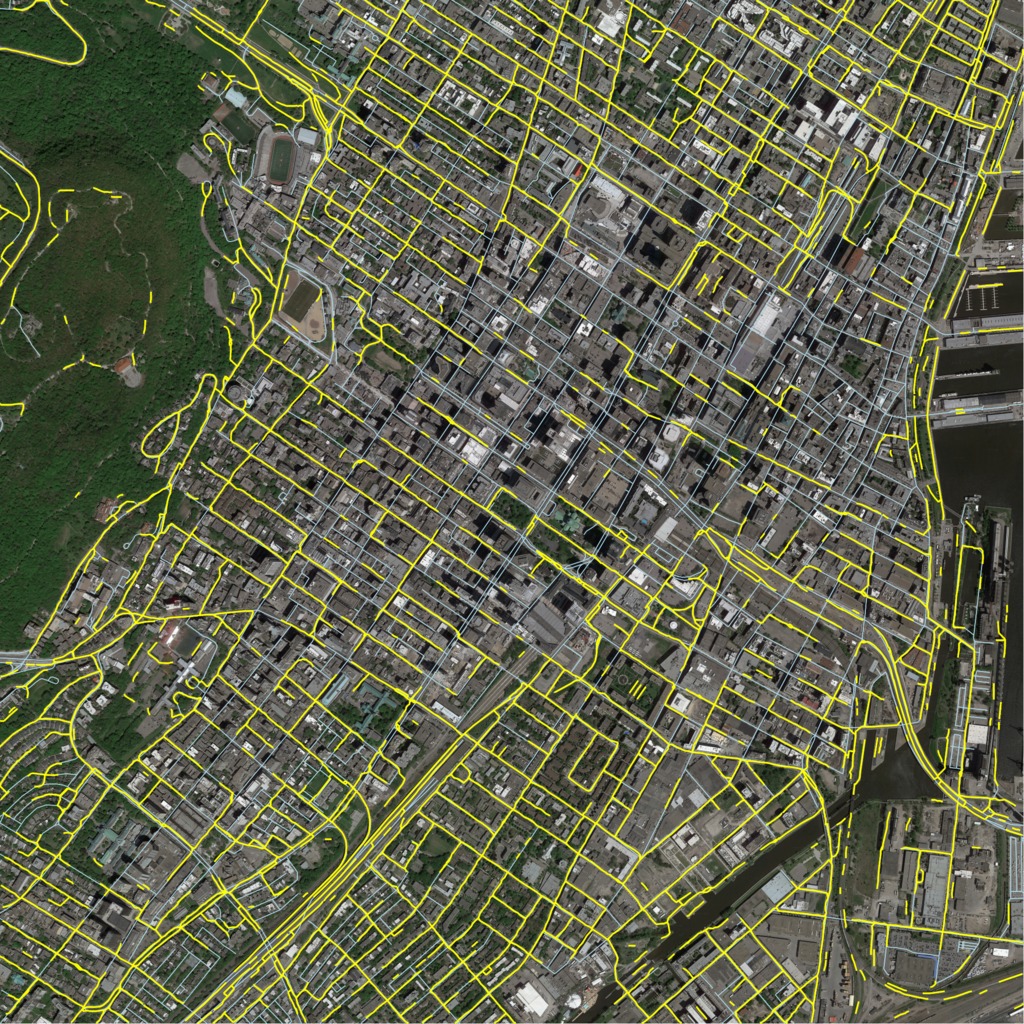}\hfill \\
    
    \caption{More qualitative results of our method (left) compared to RoadTracer \cite{Bastani2018} (middle) and DeepRoadMapper \cite{Mattyus2017} (right), on the RoadTracer test set.}
    \label{fig:quali1}
\end{figure*}

\begin{figure*}[p!]
    \centering
    \includegraphics[trim={2mm 2mm 2mm 2mm},width=0.32\textwidth]{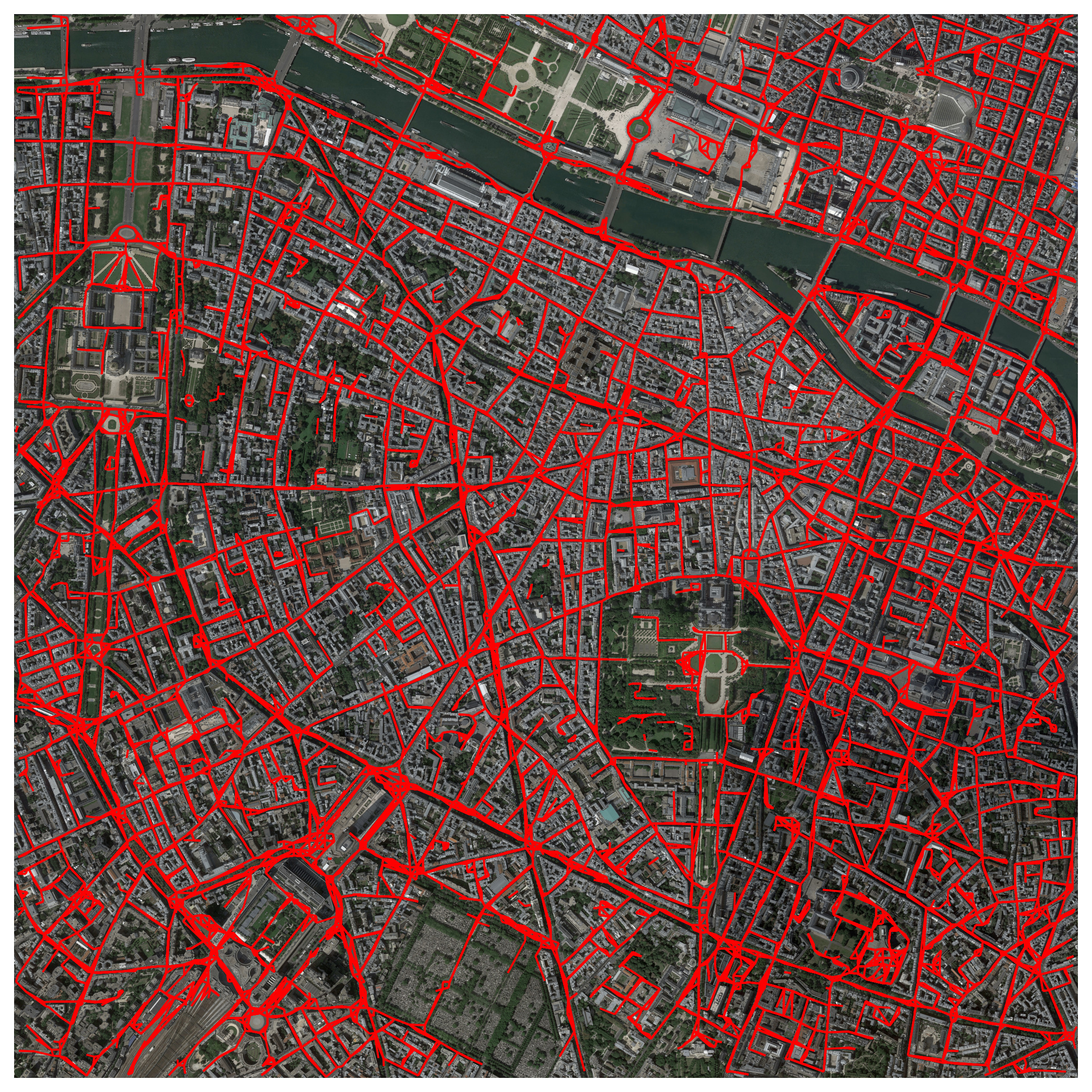}
    \includegraphics[width=0.32\textwidth]{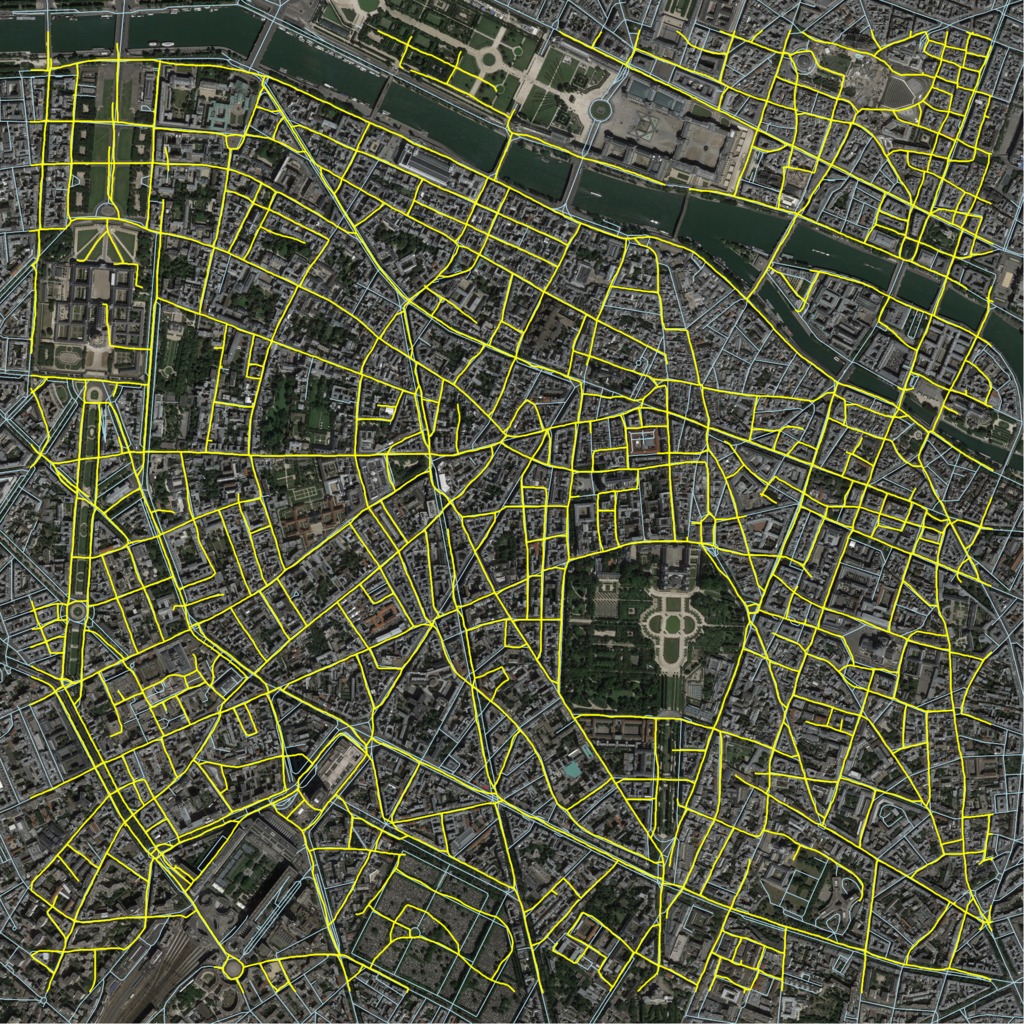}
    \includegraphics[width=0.32\textwidth]{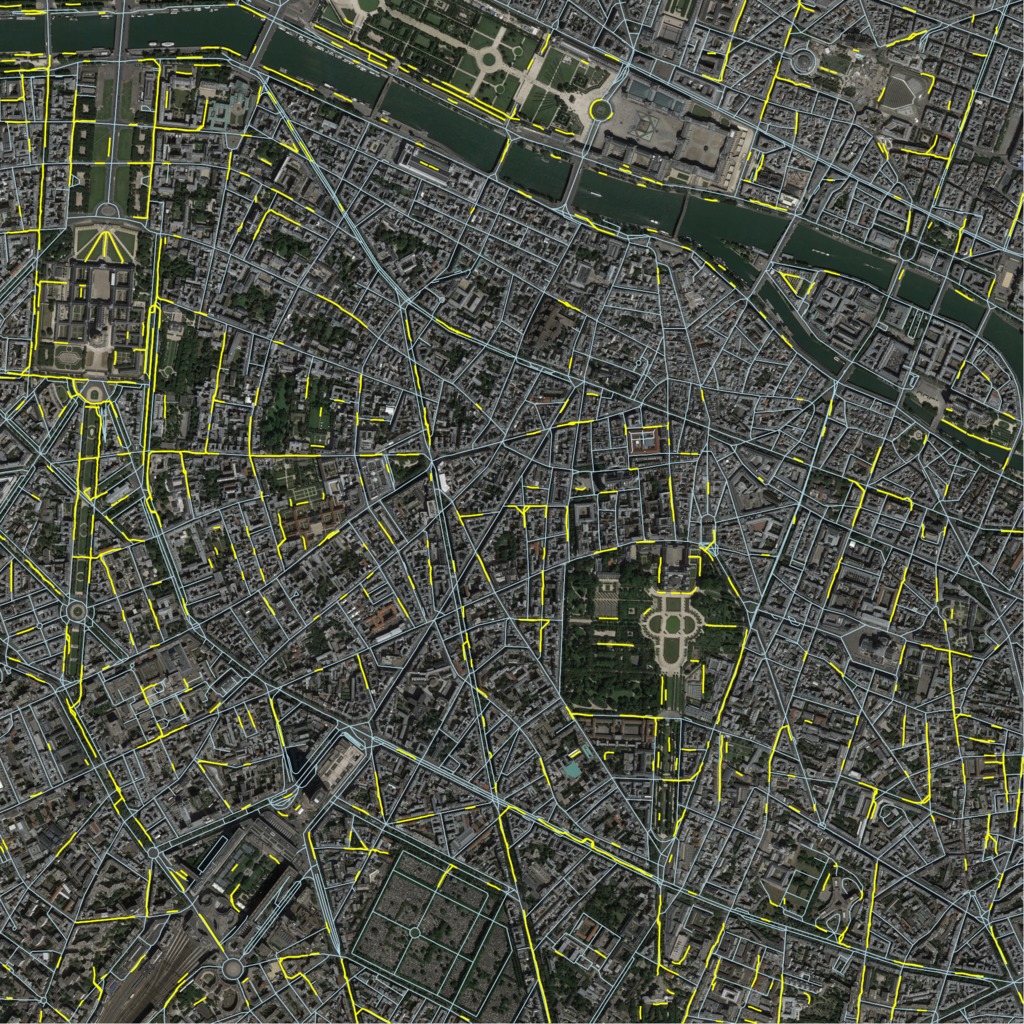}\hfill \\
    
    \includegraphics[trim={2mm 2mm 2mm 2mm},width=0.32\textwidth]{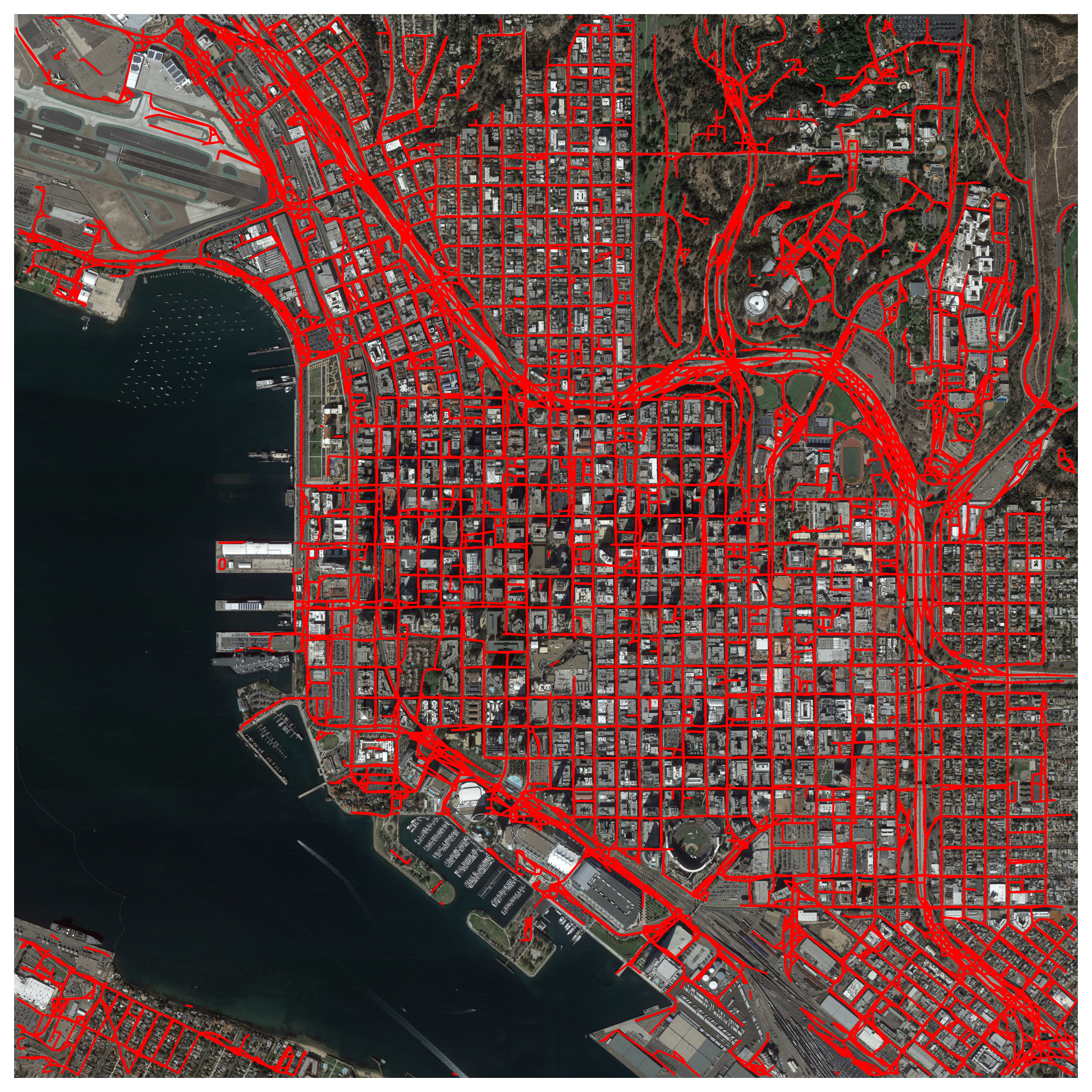}
    \includegraphics[width=0.32\textwidth]{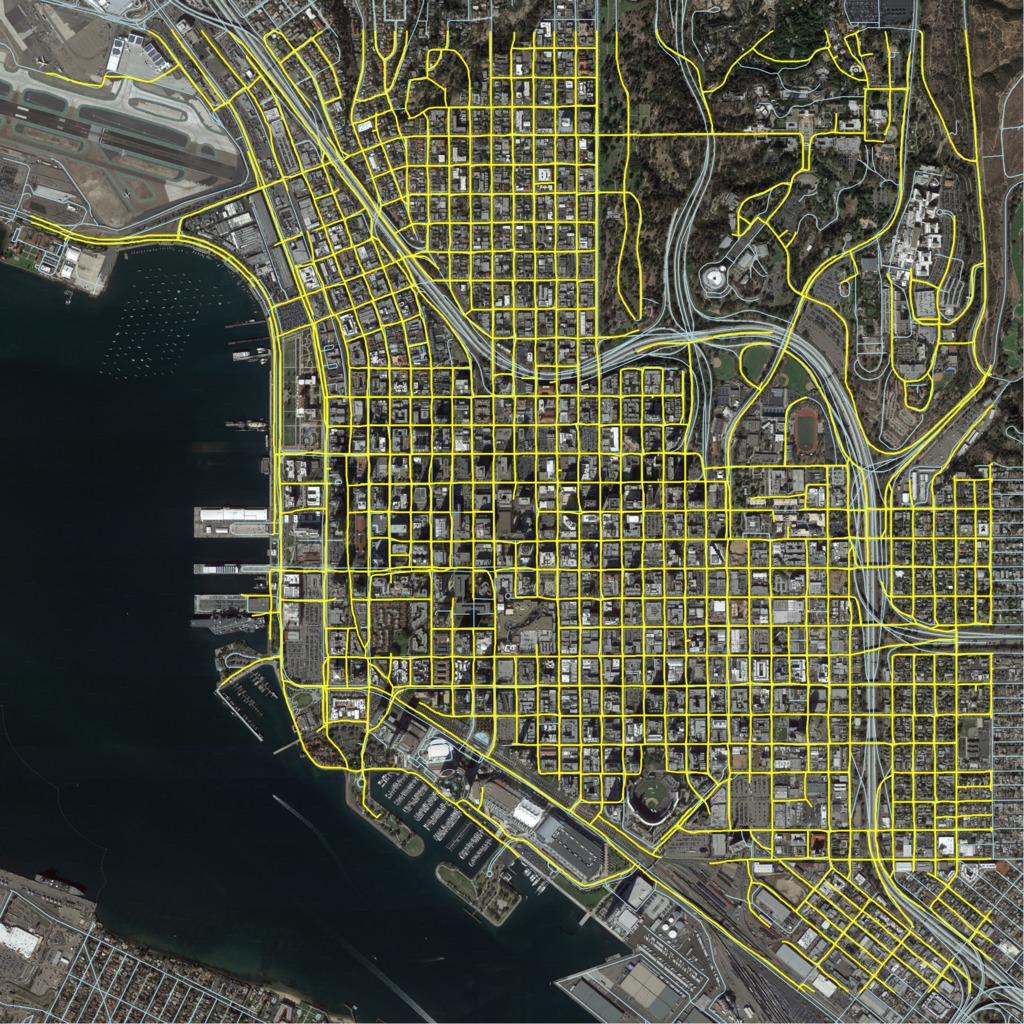}
    \includegraphics[width=0.32\textwidth]{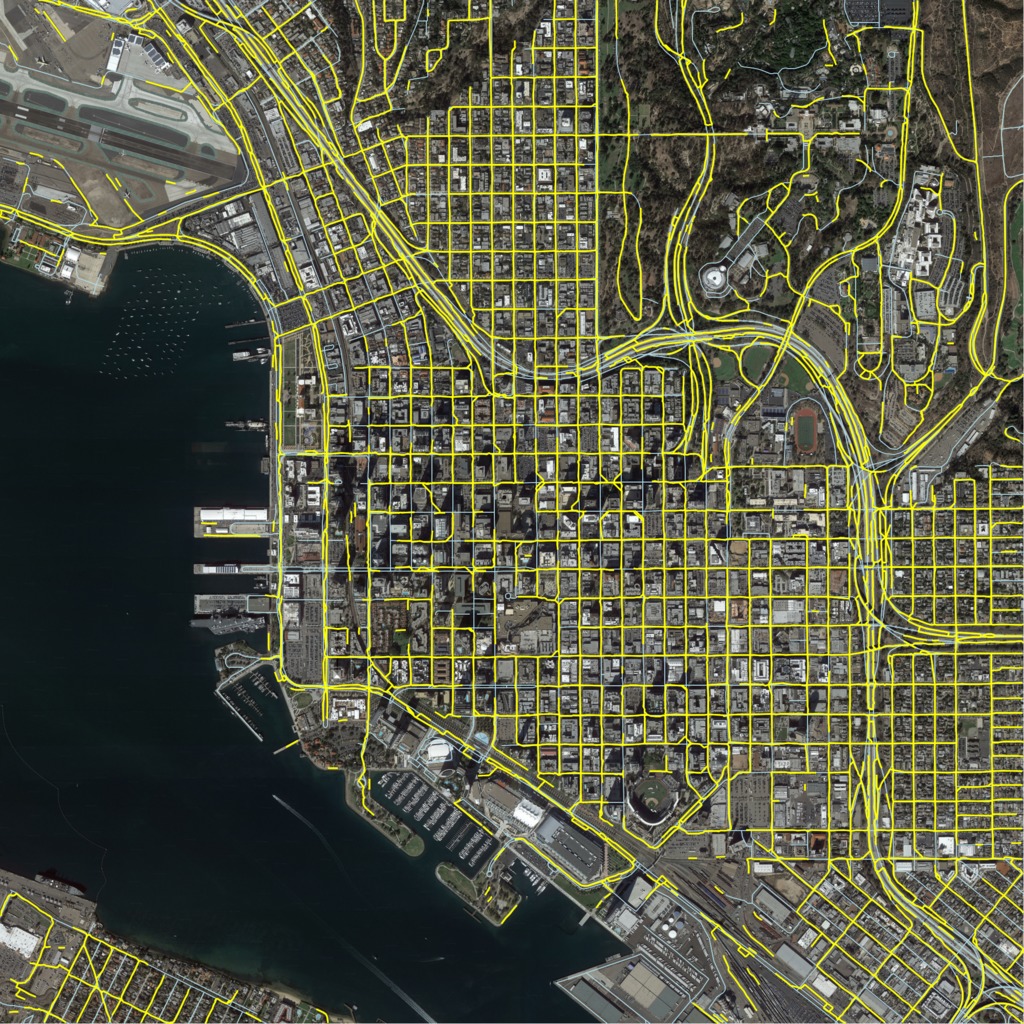}\hfill \\
    
    \includegraphics[trim={2mm 2mm 2mm 2mm},width=0.32\textwidth]{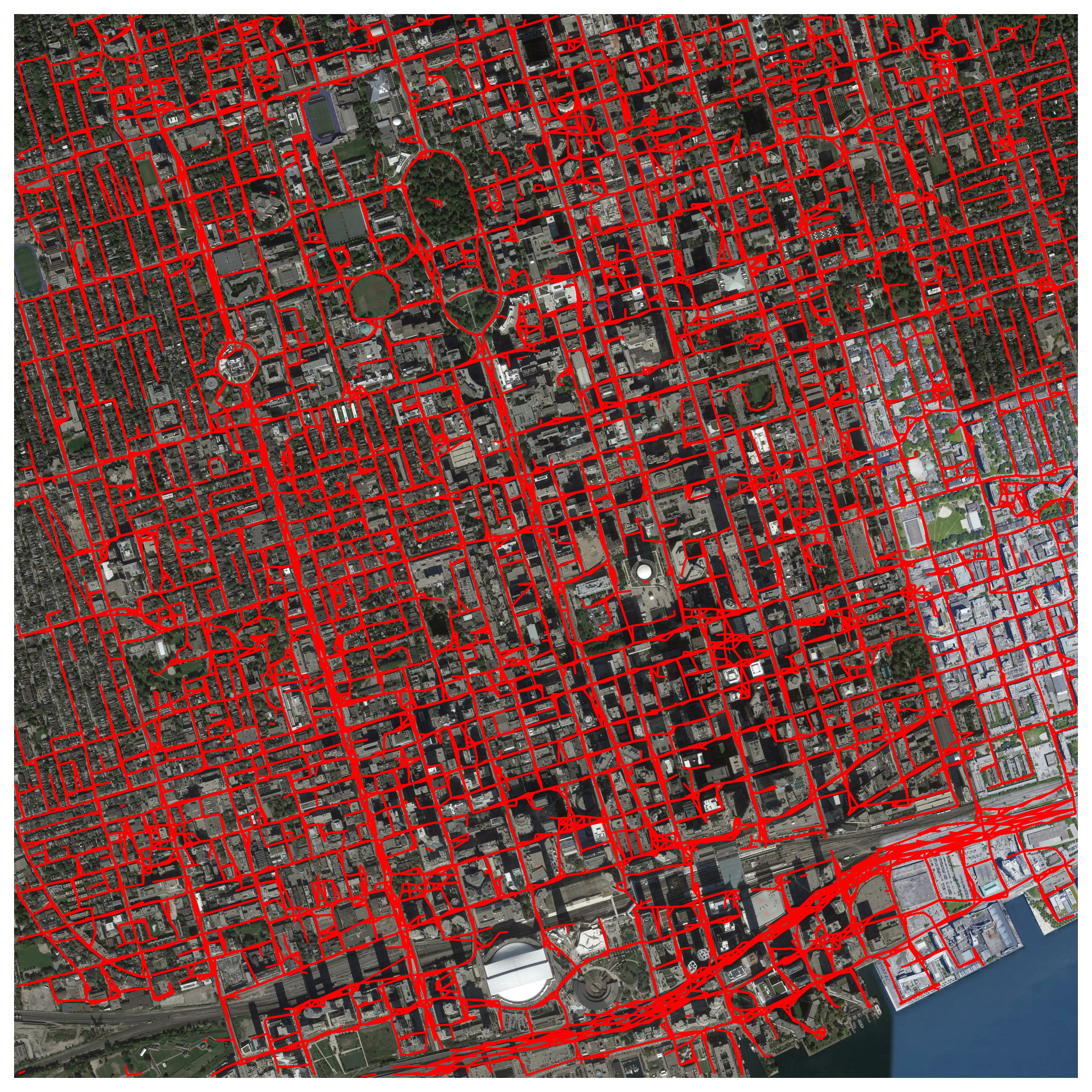}
    \includegraphics[width=0.32\textwidth]{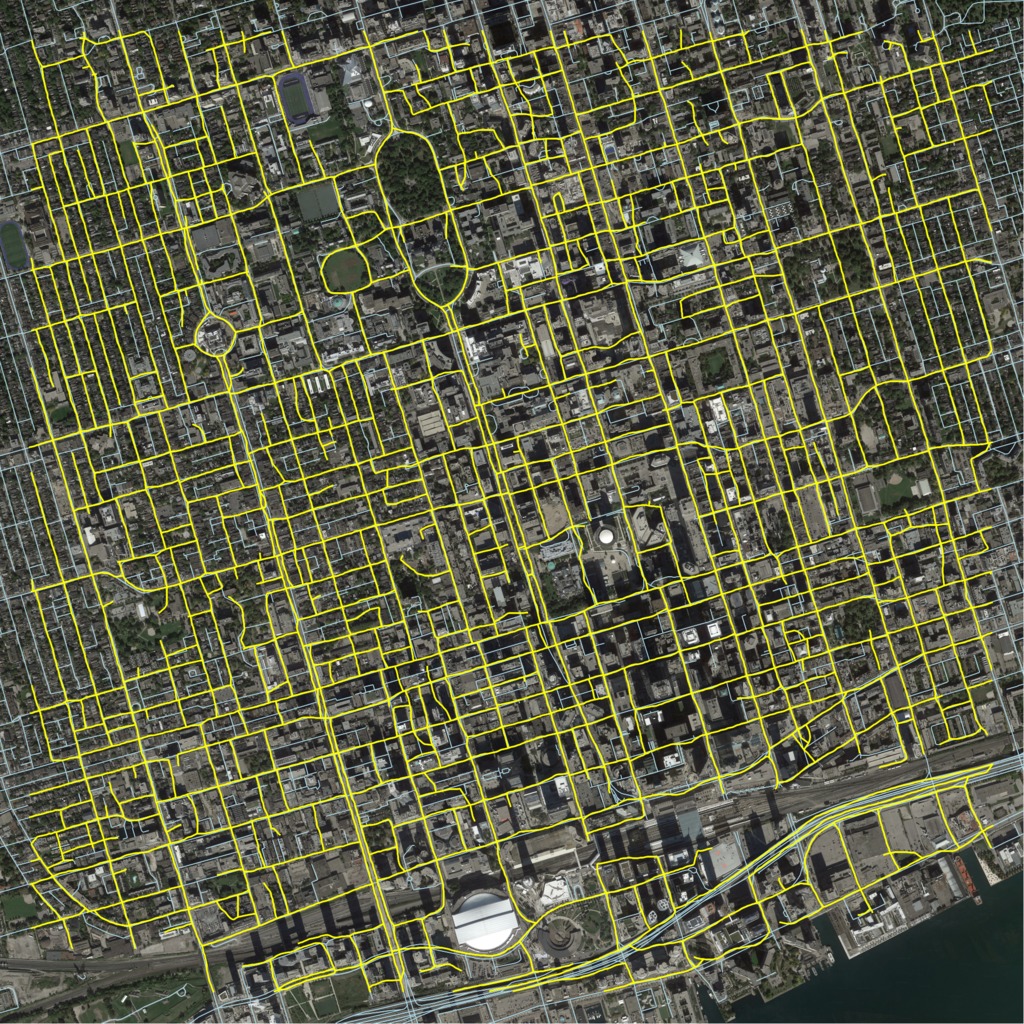}
    \includegraphics[width=0.32\textwidth]{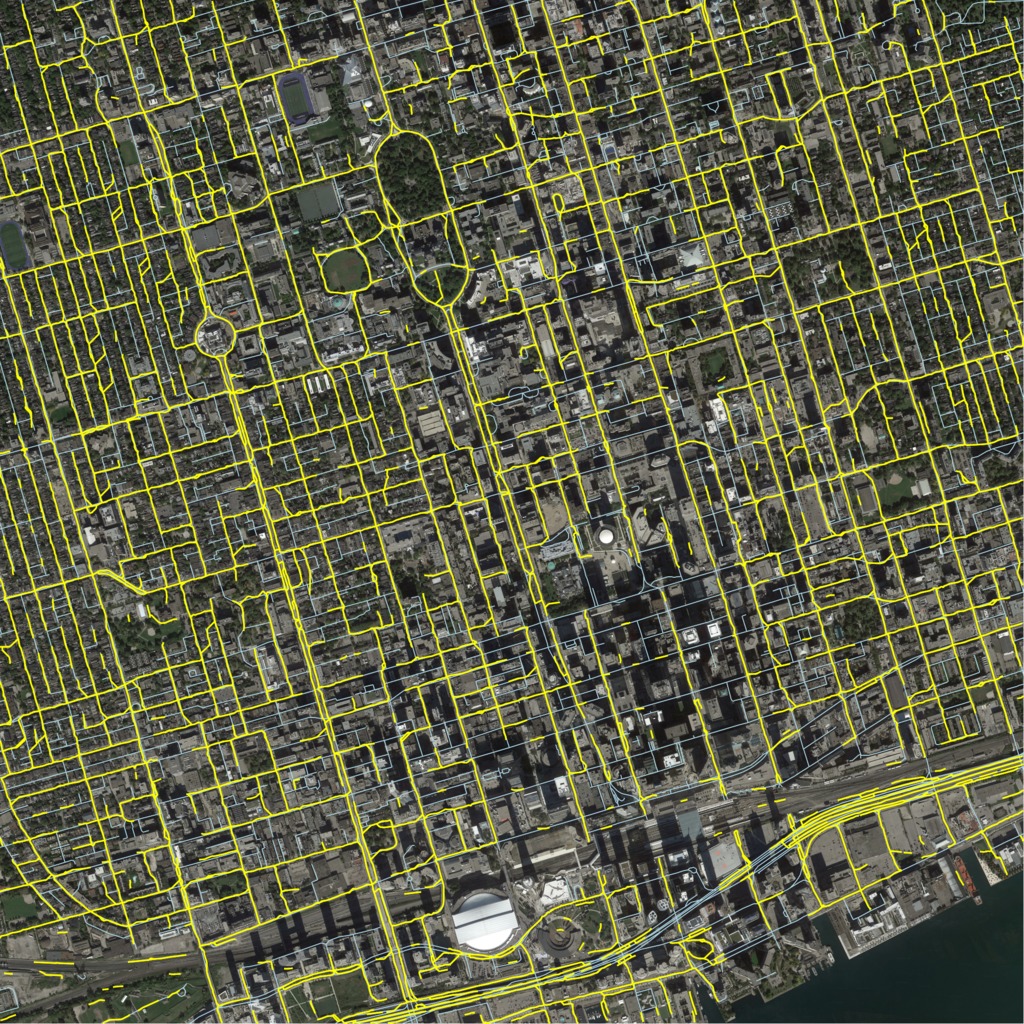}\hfill \\
    
    \caption{More qualitative results of our method (left) compared to RoadTracer \cite{Bastani2018} (middle) and DeepRoadMapper \cite{Mattyus2017} (right), on the RoadTracer test set.}
    \label{fig:quali2}
\end{figure*}

Figures \ref{fig:quali1} and \ref{fig:quali2} show more qualitative results
on cities from the RoadTracer test set. Our method is able to find more roads
than RoadTracer \cite{Bastani2018} and DeepRoadMapper \cite{Mattyus2017}. Some roads and highways that cannot be accessed easily
by iterative approaches are found by our method. Our graphs also seem to have a better
connectivity than the ones found by the segmentation-based method.

\section{Environmental impact statement}

While the focus of our work is to create efficient neural networks 
that are able to run on very low power devices, we cannot help but 
notice that these networks are still created and trained using
power-hungry multi-GPU machines and wonder about the environmental 
impact. For this paper, we tracked the number of GPU hours used
and use that to estimate its global environmental footprint and publish these estimations, as recommended in \cite{henderson2020towards,anthony2020carbontracker,lannelongue2020green}.

In order to run the experiments required for our main results and 
ablation study, we have used 1572 GPU hours on Nvidia Tesla V100
GPUs, which are rated for a power consumption of 300W. This, not counting
CPUs, cooling, PSU efficiency, storage of datasets and results,
as well as different trials or hyper-parameter searches on workstations,
amounts to 471,6kWh. Since the carbon intensity of our electricity grid is 10
gCO2/kWh, we estimate an emission of 4716 gCO2, which is equivalent to 39.17 km traveled by car according to \cite{anthony2020carbontracker}.

\clearpage

{\small
\bibliographystyle{ieee_fullname}
\bibliography{biblio}
}

\end{document}